\documentclass{article}

\PassOptionsToPackage{numbers, compress}{natbib}
\usepackage[preprint]{neurips_2025}

\usepackage[utf8]{inputenc} % allow utf-8 input
\usepackage[T1]{fontenc}    % use 8-bit T1 fonts
\usepackage{url}            % simple URL typesetting
\usepackage{booktabs}       % professional-quality tables
\usepackage{amsfonts}       % blackboard math symbols
\usepackage{nicefrac}       % compact symbols for 1/2, etc.
\usepackage{microtype}      % microtypography
\usepackage{xcolor}         % colors
\usepackage{amsmath}
\usepackage{amssymb}
\usepackage{bm}

\usepackage{lineno}
\usepackage{newfloat}
\usepackage[marginal]{footmisc}
\usepackage{framed,multirow}
\usepackage{makecell}
\usepackage{threeparttable}
\usepackage{colortbl}
\usepackage{bbding}
\usepackage{color}
\usepackage{adjustbox}
\usepackage{caption}
\usepackage{float}
\usepackage{amsmath,amsfonts,bm}

\def\eqref#1{Eqn.~\ref{#1}}
\def\1{\bm{1}}

\def\vr{{\bm{r}}}
\def\vs{{\bm{s}}}

\def\vx{{\bm{x}}}

\DeclareMathAlphabet{\mathsfit}{\encodingdefault}{\sfdefault}{m}{sl}
\SetMathAlphabet{\mathsfit}{bold}{\encodingdefault}{\sfdefault}{bx}{n}

\renewcommand{\frac}{\dfrac}
\def\cite#1{\citep{#1}}

\usepackage{breakurl}
\usepackage[breaklinks]{hyperref}
\definecolor{darkblue}{rgb}{0.0,0.0,0.65}

\hypersetup{
  colorlinks = true,
  citecolor  = darkblue,
  linkcolor  = red,
  citecolor  = darkblue,
  filecolor  = darkblue,
  urlcolor   = darkblue,
}

\title{Multi-Modal Explainable Medical AI Assistant for Trustworthy Human-AI Collaboration}

\newcommand{\methodname}{XMedGPT}

\author{%
\textbf{Honglong Yang}$^{1}$,
\textbf{Shanshan Song}$^{1}$,
\textbf{Yi Qin}$^{1}$,
\textbf{Lehan Wang}$^{1}$,
\textbf{Haonan Wang}$^{1}$, \\
\textbf{Xinpeng Ding}$^{1}$,
\textbf{Qixiang Zhang}$^{1}$,
\textbf{Bodong Du}$^{1}$,
\textbf{Xiaomeng Li}$^{1,2}$\thanks{Corresponding author: Xiaomeng Li (eexmli@ust.hk).} \\
$^1$Department of Electronic and Computer Engineering, \\
The Hong Kong University of Science and Technology \\
$^2$Department of Computer Science and Engineering, \\
The Hong Kong University of Science and Technology \\
}

\usepackage{hyperref}       % hyperlinks
\usepackage{cleveref}

\begin{document}

\maketitle

\begin{abstract}

Generalist Medical AI (GMAI) systems have demonstrated expert-level performance in biomedical perception tasks, yet their clinical utility remains limited by inadequate multi-modal explainability and suboptimal prognostic capabilities. Here, we present \methodname{}, a clinician-centric, multi-modal AI assistant that integrates textual and visual interpretability to support transparent and trustworthy medical decision-making. \methodname{} not only produces accurate diagnostic and descriptive outputs, but also grounds referenced anatomical sites within medical images, bridging critical gaps in interpretability and enhancing clinician usability. To mitigate overreliance and support real-world deployment, we introduce a reliability indexing mechanism that quantifies uncertainty through consistency-based assessment via interactive question-answering. We validate \methodname{} across four pillars: multi-modal interpretability, uncertainty quantification, and prognostic modeling, and rigorous benchmarking. The model achieves an Intersection over Union (IoU) of 0.703 across 141 anatomical regions, and a Kendall’s tau-b of 0.479 ($P$ < 0.05), demonstrating strong alignment between visual rationales and clinical outcomes. For uncertainty estimation, it attains an AUC of 0.862 on visual question answering and 0.764 on radiology report generation. Trained on over 7 million image-text pairs—including 1.6 million with pixel-level annotations—\methodname{} enables precise tumor morphology analysis and longitudinal change detection. In survival and recurrence prediction for lung and glioma cancers, it surpasses prior leading models by 26.9\%, and outperforms GPT-4o by 25.0\%. Rigorous benchmarking across 347 datasets covers 40 imaging modalities and external validation sans four anatomical systems confirming exceptional generalizability, with performance gains surpassing existing GMAI by 20.7\% for in-domain evaluation and 16.7\% on 11,530 in-house data evaluation. Together, XMedGPT represents a significant leap forward in clinician-centric AI integration, offering trustworthy and scalable support for diverse healthcare applications.

\end{abstract}

\section*{Introduction}

\begin{figure}[!b]
    \centering
    \includegraphics[width=1.0\textwidth]{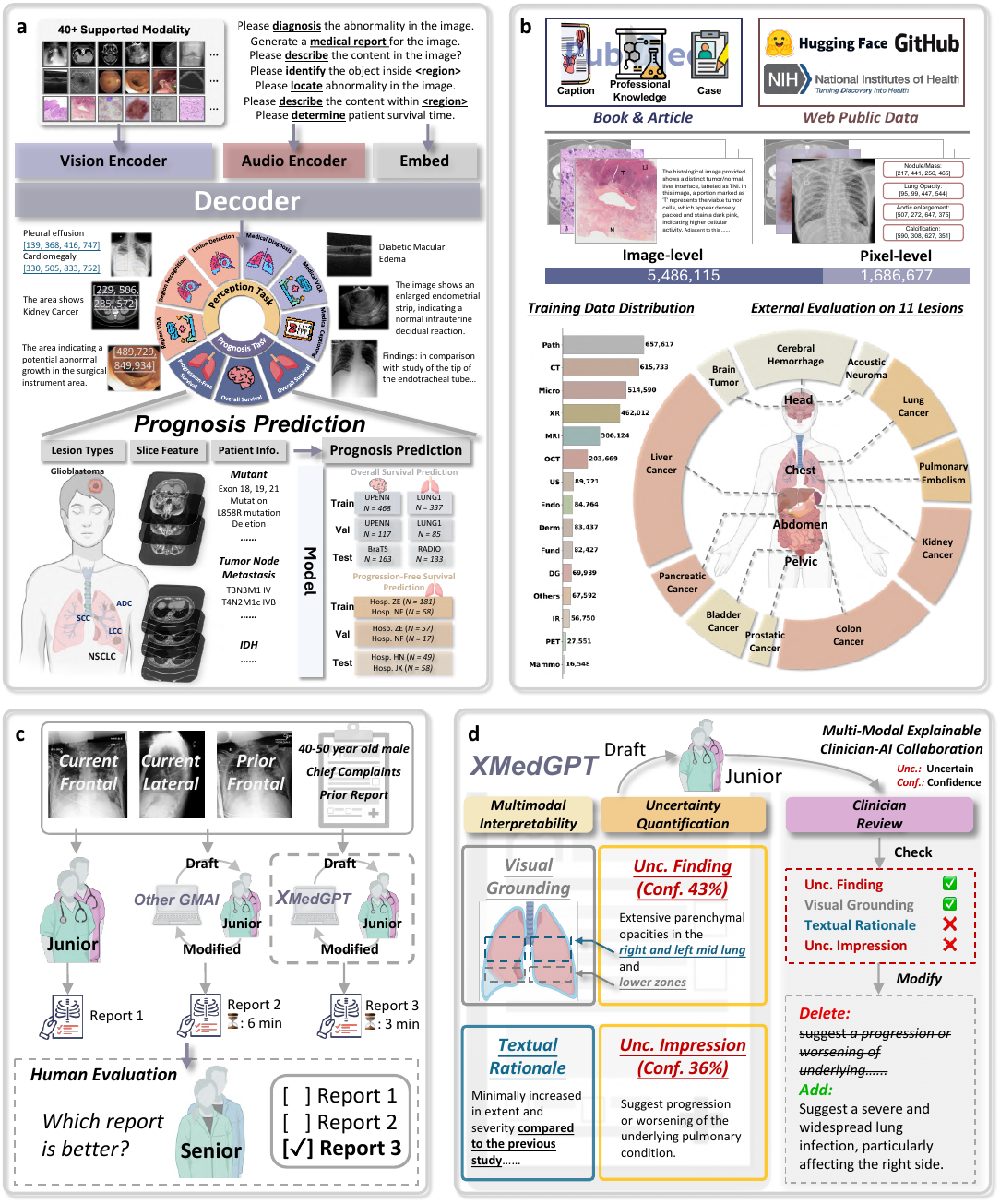}
\end{figure}
\begin{figure}[!t]
\centering
\caption{\textbf{a,} \methodname{} is a versatile multi-modal AI framework designed to handle over 40 distinct image modalities while seamlessly integrating inputs from images, text, audio. Beyond image-level perception tasks, it excels in pixel-level analysis alongside complex prognostic modeling, including overall and progression-free survival predictions. \textbf{b,} The training dataset consists of 7 million image-text, including 1.6 million with pixel-level annotations. External validation spans 11 lesion types across 4 anatomical regions, including the brain, lungs, abdomen, and pelvis. \textbf{c,} In the human-centric evaluation, senior clinicians blindly assessed three reports from (1) human-written, (2) junior clinician + baseline AI (GMAI), and (3) junior clinician + \methodname{}. Senior clinicians provided detailed feedback to identify the best approach. \textbf{d,} \methodname{}-clinician collaboration flow. The model drafts a report with multi-modal interpretability and a reliability index, which the clinician refines by verifying the provided evidence. This process enhances clinical decision-making by combining AI’s analytical power with clinician expertise.}
\label{fig:intro}
\end{figure}

Multimodal Large Language Models (MLLMs), trained on extensive and diverse datasets, have driven transformative advances in artificial intelligence~\cite{hurst2024gpt4o,liu2024deepseekv3,wang2024qwen2vl,liu2023visual_llava,chen2024internvl} and catalyzed the development of Generalist Medical AI (GMAI) in healthcare.
Unlike domain-specific foundation models~\cite{lu2024conch,xiang2025vision_MUSK,chen2024towards_CPath,zhou2023foundation_RETFound,kim2024transparent_monet,christensen2024vision_EchoCLIP,pai2024foundation_fmcib} and pre-trained biomedical models~\cite{zhang2023biomedgpt,zhang2025multimodal_biomedclip} that require extensive task-specific adaptation to perform downstream tasks, GMAI~\cite{zhang2023biomedgpt,moor2023medflamingo,wu2023radfm,li2024llavamed,he2024meddr,wang2024medrega} represents a multimodal, task-versatile system capable of integrating diverse medical modalities and generating outputs (e.g., interactive clinical guidance) to optimize healthcare delivery and enhance clinical decision-making. While MLLMs underpin these systems by integrating visual and textual biomedical data, their translation into trustworthy, clinician-ready tools remains insufficiently explored, primarily due to the following unresolved challenges.

First, existing GMAI systems lack multimodal explainability, which undermines their trustworthiness for use in clinical practice. Although these models can generate text-based outputs for diverse inputs, including medical images and clinician instructions, they may produce unreliable diagnostic conclusions without visual explainability pinpointing the precise locations of potential diseases and textual explainability to articulate diagnostic reasoning and quantify prediction confidence. This deficiency in multimodal interpretability can lead clinicians to inadvertently rely on unverified outputs without thoroughly cross-referencing them with the medical images and patient information, leading to potential misdiagnosis~\cite{kostick2025caution, comeau2025preventing, jongsma2024we, rajpurkar2022ai, rao2025multimodal}. Second, existing GMAI systems are mainly limited to performing perception tasks, achieving performance comparable to that of clinicians. However, they lack the ability to learn from relationships among multiple medical images and to model complex dynamics, such as those involving multiple visits or longitudinal patient data. Consequently, these systems remain limited to perform advanced prognostic tasks, such as predicting treatment outcomes (e.g., survival or recurrence) which require integrating multi-source information like patient demographics, imaging biomarkers, and longitudinal disease progression. These tasks are challenging even for expert clinicians, who often struggle to consistently and accurately quantify patient risks~\cite{hager2024evaluation, fathi2025multiparametric, xiang2025vision_MUSK, acosta2022multimodal}. Third, prior work lacks rigorous external evaluation using real-world clinical data and is often limited to institution-specific tasks or narrow healthcare contexts.

Here, we propose \methodname{}, a clinician-centric, multi-modal explainable medical AI assistant that employs a unified framework to seamlessly integrates visual and textual explainability. \methodname{} not only generates accurate text outputs but also visually highlights the anatomical locations referenced in the text within medical images, effectively addressing critical gaps in interpretability and significantly enhancing clinical usability. To foster trustworthy human-AI collaboration, we introduce the Reliability Indexing mechanism, which assesses model uncertainty by evaluating the consistency of the system's predictions through interactive question-and-answer interactions.
Instead of relying solely on the initial output, which may be influenced by the model’s learned reliance on common reporting templates or structural conventions, \methodname{} breaks down the output into a series of related visual question-answering (VQA) pairs—addressing questions about the presence, location, and severity of lesions.
The critical attributes in the sentence are then validated using these image-driven VQA responses. The reliability index is computed by measuring the entropy of the consistency distribution between the attributes extracted from the sentence and the corresponding image-based answers, classifying the results into three reliability levels—high, medium, or low. This approach allows clinicians to efficiently identify areas requiring further review, enhancing the trustworthiness and interpretability of the model's outputs and fostering improved human-AI collaboration.

Our \methodname{} was trained on a large-scale biomedical image dataset, which includes over 7 million image-text pairs, including 1.6 million pairs with pixel-level annotations, spanning 40 medical modalities and 141 anatomical regions. This extensive training enables \methodname{} to conduct pixel-level analysis of tumor morphology and detect subtle longitudinal changes across multiple visits — critical factors for accurate prognostic modeling. We demonstrate that \methodname{} unlocks state-of-the-art accuracy in predicting critical outcomes such as survival and recurrence (\Cref{fig:intro}a) from high-incidence cancers (e.g., lung adenocarcinoma) to high-mortality diseases (e.g., glioblastoma). In predicting lung cancer progression-free survival (PFS), which requires integration of patient data from multiple visits, \methodname{} achieved an AUC of 0.725 (95\% CI: 0.635–0.809). In comparison, previous leading model~\cite{deng2022deep_ESBP}  attained an AUC of 0.625 (95\% CI: 0.511–0.727), and GPT-4o~\cite{hurst2024gpt4o} achieved an AUC of 0.620 (95\% CI: 0.526–0.712). Additionally, for survival prediction in glioblastoma and non-small cell lung cancer (NSCLC) using single-visit data, \methodname{} consistently outperforms previous state-of-the-art model~\cite{pai2024foundation_fmcib,chen2019med3d} by 16.9\% and GPT-4o~\cite{hurst2024gpt4o} by 29.05\%. On 23 public benchmarks, \methodname{} consistently outperformed previous state-of-the-art (SOTA) GMAI model~\cite{wang2024medrega} by 26.2\% and surpassed GPT-4o by 29.1\% across 18 biomedical tasks while achieving a 29.3\% F1-score improvement in pixel-level lesion characterization over prior SOTA model~\cite{huang2024bird}. In a rigorous external validation using real-world clinical data covering nine common cancers across four anatomical regions (head, chest, abdomen, and pelvis) and two high-incidence lesions (cerebral hemorrhage and pulmonary embolism), \methodname{} demonstrated robust generalizability, outperforming the top-performing GMAI model~\cite{wang2024medrega} by 16.7\%.

Notably, the effectiveness of clinician-centric multimodal interpretability was evaluated across three
key pillars. Specifically, \methodname{} demonstrated superior grounding performance, achieving an
Intersection over Union (IoU) score of 0.703 across 141 anatomies and a Kendall’s tau-b correlation
coefficient of 0.479 ($P$ < 0.05), indicating strong alignment between rationales and clinical outcomes.
For uncertainty quantification, the model also achieved an AUC of 0.862 on visual question-answering
(VQA) tasks and a 0.764 AUC on report generation. These results underscore the clinical viability
of fusing trustworthy AI systems into routine clinical practice, elevating diagnostic precision while
safeguarding against diagnostic cascades triggered by opaque AI errors. In summary, \methodname{}
represents a significant leap forward in the integration of AI into healthcare, combining multi-modal
data synthesis, interpretability, prognostic precision, and real-world validation to deliver a robust,
clinician-ready decision-support tool with broad clinical applicability.

% Notably, in prospective human-centric evaluations simulating real-world practice (\Cref{fig:intro}c), a blinded expert assessment by a panel of senior board-certified radiologists demonstrated that 72\% of reports generated through \methodname{}-clinician collaboration were rated as superior or equivalent to those written solely by junior clinicians. Crucially, \methodname{}-clinician collaboration alleviates risks of clinician over-reliance on opaque AI outputs through a 15.2\% reductions of AI-misleading errors compared to conventional GMAI-assisted workflows. These findings underscore the clinical viability of fusing trustworthy AI systems into routine clinical practice, elevating diagnostic precision while safeguarding against diagnostic cascades triggered by opaque AI errors. In summary, \methodname{} represents a significant leap forward in the integration of AI into healthcare, combining multi-modal data synthesis, interpretability, prognostic precision, and real-world validation to deliver a robust, clinician-ready decision-support tool with broad clinical applicability.

\section*{Results}

\subsection*{\methodname{} Achieves Superior Performance Across 23 Benchmarks Beyond 38 Imaging Modalities}

To rigorously benchmark \methodname{} against leading GMAI models, we conduct a comprehensive evaluation across five core tasks: single-label diagnosis, multi-label diagnosis, visual-question answering (VQA), multiple-choice reasoning, and image captioning. For single-label diagnosis, \methodname{} is tested on nine datasets spanning organ recognition (OrganAMNIST, OrganCMNIST, OrganSMNIST)~\cite{yang2023medmnist} and lesion classification (PathMNIST, OCTMNIST, PneumoniaMNIST, BreastMNIST, PAD-UFES-20~\cite{pacheco2020padufes20}, MURA~\cite{rajpurkar2017mura}), while multi-label diagnosis assesses its performance on diverse conditions including spinal (VinDr-SpineXR~\cite{nguyen2021vindrspinexr}), thoracic (VinDr-CXR~\cite{nguyen2022vindrcxr}, VinDr-PCXR~\cite{pham2022vindrpcxr}, VinDr-Mammo~\cite{nguyen2023vindrmammo}, ChestMNIST~\cite{yang2023medmnist}), and ocular (RFMiD~\cite{panchal2023rfmid}, BRSET~\cite{nakayama2024brset}) disorders. The VQA evaluation covers radiology (VQA-Rad~\cite{vqarad}, SLAKE~\cite{liu2021slake}) and pathology (PathVQA~\cite{he2020pathvqa}) datasets, probing anatomical and tissue-level understanding. To stress-test generalization, we leverage two large-scale multiple-choice benchmarks: OmnimedVQA (42 datasets, 8 modalities)~\cite{hu2024omnimedvqa} and GMAI-MMbench (284 datasets, 38 modalities, 18 tasks)~\cite{ye2024gmaimmbench}, which surpass the scope of prior GMAI assessments. Finally, image captioning is evaluated on structured CXR datasets (IU-Xray~\cite{demner2016iuxray}, MIMIC-CXR~\cite{johnson2019mimiccxr}). This unified evaluation across tasks, modalities, and granularities not only demonstrates \methodname{}'s versatility but also establishes its superiority over existing GMAI models in both depth and breadth of medical understanding. We employ Macro-F1 as the primary metric for both single-label and multi-label diagnosis tasks, ensuring balanced evaluation across all classes. For VQA, we follow community standards by using F1-score as our main metric, supplemented by closed-ended accuracy (Closed-Acc) and open-ended recall (Open-Recall) for comprehensive analysis (see \Cref{tab:vqa}). Multiple-choice tasks are evaluated using answer selection accuracy. In assessing image captioning quality, we utilize both linguistic metrics (ROUGE-L, BLEU, METEOR) for text fluency and CheXpert- F1 for clinical precision, providing a dual perspective on generated captions.

\begin{figure}[!b]
    \centering
    \includegraphics[width=1.0\linewidth]{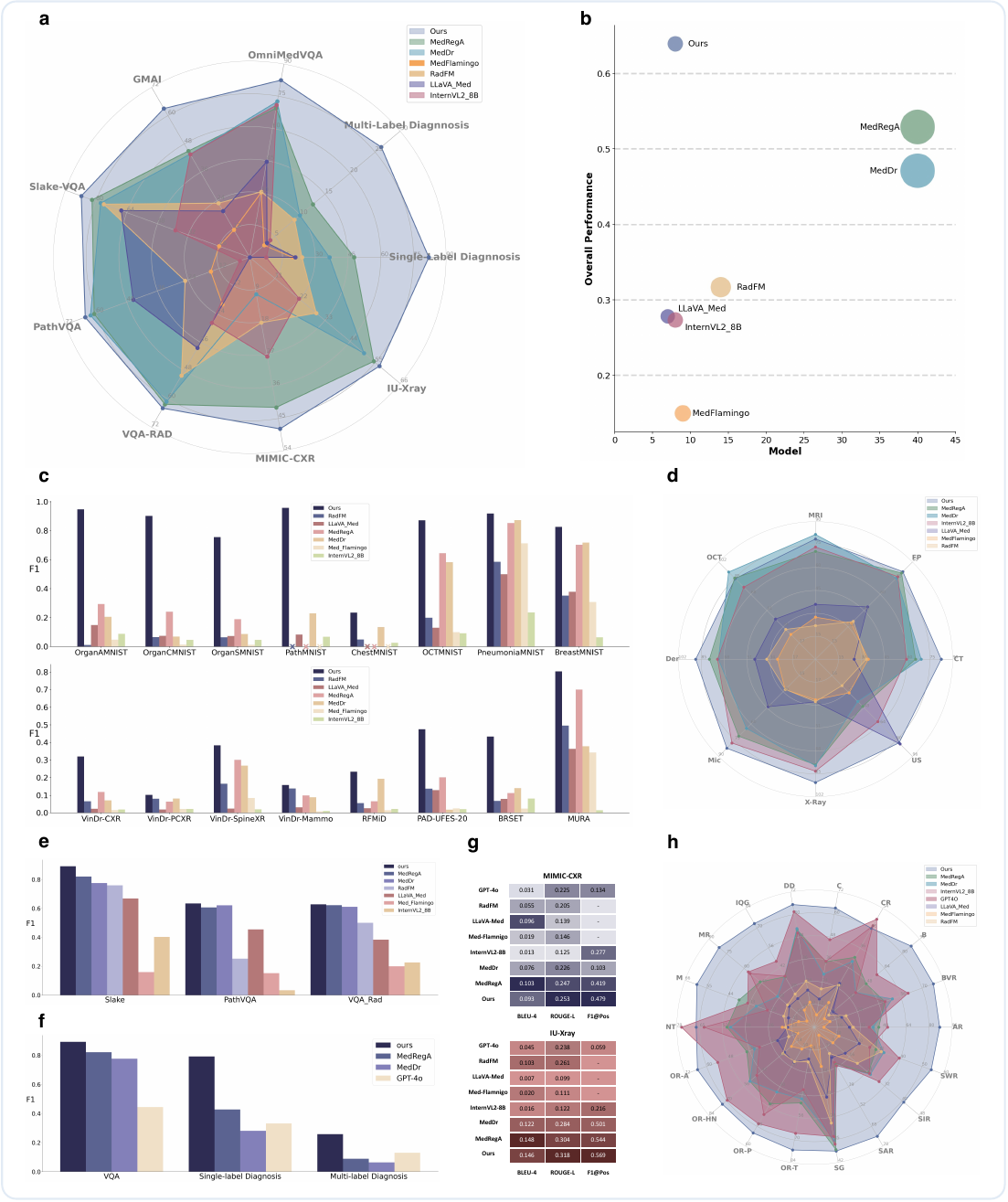}
    \caption{\textbf{a,} \methodname{} consistently outperforms other generalist biomedical models across five core tasks: single-label diagnosis, multi-label diagnosis, visual question answering (VQA), multiple-choice reasoning, and image captioning. \textbf{b,} Overall task performance relative to model scale. \textbf{c,} Performance on single- and multi-label diagnosis tasks, evaluated using F1-score and micro-F1-score, respectively. \textbf{d,} Accuracy on the OmniMedVQA multiple-choice benchmark, covering eight medical modalities from 42 datasets. \textbf{e,} Comparison of closed-ended accuracy, F1-score, and open-ended recall for medical VQA; 'x' indicates missing results in the original publications. \textbf{f,} Benchmarking against GPT-4o and other leading GMAI models on VQA, single-label, and multi-label diagnosis tasks. \textbf{g,} Image captioning performance on MIMIC-CXR and IU-Xray, assessed using BLEU, ROUGE-L, and CheXpert-F1 metrics. \textbf{h,} Accuracy on the GMAI-MMbench multiple-choice benchmark, covering 284 datasets, 38 modalities, and 18 clinical tasks.}
    \label{fig:page2}
\end{figure}

As shown in \Cref{fig:page2}, \methodname{} achieves a unified F1 score of 0.639 across all five tasks, surpassing the previous state-of-the-art (SOTA) of 0.529 by MedRegA~\cite{wang2024medrega} by a significant margin. Notably, this performance is achieved with only 8 billion parameters, far fewer than the 40B parameters of MedRegA and MedDr~\cite{he2024meddr}. For single-label diagnosis (See \Cref{tab:sinlge_multi_label_diagnosis} for detail), \methodname{} attains a Macro-F1 score of 0.821, outperforming MedRegA 0.479 by a large margin. On the challenging long-tailed multi-label diagnosis task (See \Cref{tab:sinlge_multi_label_diagnosis} for detail), \methodname{} achieves a Macro-F1 score of 0.262 across seven datasets, vastly exceeding prior models (ranging from 0.028–0.126). In addition, \methodname{} sets new SOTA results on major VQA benchmarks: 0.891 F1 on SLAKE, 0.634 on PathVQA, and 0.628 on VQA-RAD, outperforming 40B-parameter models MedDr and MedRegA (See \Cref{tab:vqa} for detail). For closed-ended questions, it achieves 90.95\% Closed-Acc on SLAKE and 92.39\% on PathVQA, exceeding prior best scores of 89.90\% and 90.21\%. On VQA-RAD, \methodname{} attains 80.48\% Closed-Acc and 45.07\% Open-Recall, surpassing generalist models without fine-tuning. While LLaVA-Med~\cite{li2024llavamed} achieve higher Open-Recall with task-specific fine-tuning, \methodname{} outperforms similarly non-fine-tuned models like MedDr by 20.35\% in fair comparisons. We also evaluated \methodname{} on GMAI-MMbench and OmnimedVQA, covering diverse clinical tasks and modalities. On OmnimedVQA, our model achieved 82.44\% accuracy across 42 datasets (8 modalities), surpassing MedDr’s 72.54\%. \methodname{} set SOTA performance on 6 of 8 modalities, with notable improvements on CT (+19.0\%) and Dermoscopy (+12.9\%) (See \Cref{tab:OmniMedVQA} for detail). On GMAI-MMbench (18 clinical tasks, 284 datasets, 38 modalities), \methodname{} outperformed prior models on 17 tasks, with improvements exceeding 50\% on eight tasks, highlighting its broad medical competency (See \Cref{tab:GMAI_MMbench} for detail). Finally, for image captioning, we evaluated text similarity using NLG metrics (BLEU~\cite{papineni2002bleu}, ROUGE-L~\cite{chin2004rouge}, METEOR~\cite{banerjee2005meteor}) and clinical correctness via CheXpert-F1~\cite{smit2020chexbert}. \methodname{} achieved SOTA performance on IU-Xray and MIMIC-CXR, with CheXpert-F1 scores of 0.479 (MIMIC-CXR) and 0.569 (IU-Xray) demonstrated gains of 14.3\% and 4.6\%, underscoring its clinical accuracy (See \Cref{tab:report_generation} for detail).

In direct comparison with GPT-4o, \methodname{} establishes itself as the unequivocal leader across all five core medical tasks, outperforming both general-domain and specialized medical models. On single-label diagnosis, \methodname{} achieves a 0.791 F1-score, more than doubling GPT-4o's performance (0.331) and significantly surpassing MedDr's 0.281. For multi-label diagnosis, \methodname{} attains a 0.258 F1-score - nearly twice as high as GPT-4o's 0.131, while previous medical SOTA models MedRegA and MedDr score below 0.100. In image captioning, \methodname{} demonstrates comprehensive superiority: it exceeds GPT-4o by 33.6\% in METEOR (linguistic quality) and achieves a tenfold improvement in CheXpert-F1 (clinical accuracy). This advantage scales consistently - on the comprehensive GMAI-MMbench (284 datasets), \methodname{} outperforms GPT-4o by 29.1\% overall, with particularly striking gains exceeding 50\% on five critical clinical tasks. Together, these results underscore \methodname{}’s unique position as the first model to consistently surpass both leading general-purpose LLMs and specialized medical AI.

% \begin{figure}[!t]
%     \centering
%     \caption{\textbf{a,} \methodname{} consistently outperforms other generalist biomedical models across five core tasks: single-label diagnosis, multi-label diagnosis, visual question answering (VQA), multiple-choice reasoning, and image captioning. \textbf{b,} Overall task performance relative to model scale. \textbf{c,} Performance on single- and multi-label diagnosis tasks, evaluated using F1-score and micro-F1-score, respectively. \textbf{d,} Accuracy on the OmniMedVQA multiple-choice benchmark, covering eight medical modalities from 42 datasets. \textbf{e,} Comparison of closed-ended accuracy, F1-score, and open-ended recall for medical VQA; 'x' indicates missing results in the original publications. \textbf{f,} Benchmarking against GPT-4o and other leading GMAI models on VQA, single-label, and multi-label diagnosis tasks. \textbf{g,} Image captioning performance on MIMIC-CXR and IU-Xray, assessed using BLEU, ROUGE-L, and CheXpert-F1 metrics. \textbf{h,} Accuracy on the GMAI-MMbench multiple-choice benchmark, covering 284 datasets, 38 modalities, and 18 clinical tasks.}
%     \label{fig:page2}
% \end{figure}

\subsection*{\methodname{} is a Prognostic-Capable MLLM with Superior Performance Over Task-Specific Models}

To evaluate the potential of using GMAI model in prognostic decision-making that integrating multiple data sources, we conducted three challenging prognostic tasks: Progression-Free Survival (PFS) prediction for lung cancer patients undergoing  tyrosine kinase inhibitors (TKIs) treatment, overall survival (OS) prediction for patients with NSCLC tumors, and OS for patients with neuro-oncology (e.g., Glioma). 

\noindent\textbf{Progression-Free Survival Prediction in TKI-Treated Lung Cancer}

\noindent \methodname{} demonstrated robust performance in predicting progression-free survival (PFS) for lung cancer patients treated with tyrosine kinase inhibitors (TKIs), significantly outperforming existing methods. PFS prediction, which evaluates the time from treatment initiation to disease progression or death, is critical for risk stratification and guiding personalized therapy. To address the complexity of integrating diverse clinical data—including lung CT scans, patient demographics, medical history, disease characteristics (e.g., mutations, smoking status, metastasis), and TKI treatment parameters—we developed a model trained on a multicenter dataset from two hospitals ZE (n=238), NF (n=85) and evaluate on two external dataset from another two hospitals JX (n=58), and HN (n=49). The model was optimized to predict whether patients would achieve PFS exceeding 14 months, a clinically significant threshold based on established guidelines linking prolonged PFS to favorable treatment response and survival outcomes in TKI therapy. Evaluation on the JX and HN cohorts demonstrated strong performance of our model, achieving an AUC of 0.725 (95\% CI: 0.636–0.812), which represents a 16.7\% improvement over the leading task-specific model~\cite{deng2022deep_ESBP} (AUC: 0.621, 95\% CI: 0.511–0.729). This corresponds to a 25.7\% increase in accuracy. Kaplan-Meier analysis further confirmed the model’s clinical utility, showing statistically significant risk stratification on both JX cohort (log-rank $P$ < 0.001) and HN cohort (log-rank $P$ = 0.008), whereas the task-specific model failed to achieve significance in JX cohort ($P$ = 0.085). Similarly, when compared to the leading general-purpose model, GPT-4o, our model attained higher accuracy (0.764, 95\% CI: 0.682–0.841) and more robust risk stratification ($P$ = 0.008). In contrast, GPT-4o exhibited weaker discrimination ($P$ = 0.029) and lower accuracy (0.643, 95\% CI: 0.551–0.729), highlighting the limitations of general-purpose MLLMs in clinical settings.

\noindent\textbf{Overall Survival Estimation for NSCLC Patients}

\noindent \methodname{} demonstrated robust predictive performance in estimating overall survival (OS) for non-small cell lung cancer (NSCLC) patients. OS, defined as the duration from diagnosis to death from any cause, is a critical endpoint for risk stratification and therapeutic decision-making. To address the heterogeneity of immunotherapy response, we incorporated baseline CT imaging features, TNM staging, and histopathological subtypes into a unified predictive framework. The model was trained on the Lung1 dataset (n=420) and externally validated on the RADIO cohort (n=133), with optimization targeting 2-year survival—a clinically relevant benchmark aligned with prior studies. On independent validation, \methodname{} achieved an AUC of 0.706 (95\% CI: 0.604–0.803), representing a 18.1\% improvement over the leading clinical risk score (AUC:  0.598, 95\% CI: 0.486–0.703). Kaplan-Meier analysis revealed statistically significant stratification between risk groups ($P$ < 0.001). In contrast, the leading task-specific model's clinical risk score failed to achieve significant discrimination ($P$ = 0.125). When evaluated against GPT-4o, \methodname{} exhibited superior accuracy (0.784; 95\% CI: 0.714–0.850) compared to GPT-4o (0.699; 95\% CI: 0.624–0.782), alongside stronger risk stratification ($P$ < 0.001 for \methodname{} versus $P$ = 0.029 for GPT-4o). These results highlight the limitations of general-purpose models in oncology applications and underscore the potential of using Medical MLLMs in prognostic clinical decision-making.

\noindent\textbf{Glioblastoma Survival Estimation for Extensive Neuro-Oncology with MRI Biomarkers}

\noindent Our study further establishes the prognostic capability of \methodname{} in predicting overall survival (OS) for glioblastoma (GBM) using preoperative MRI scans, demonstrating its scalability across distinct tumor types and imaging modalities compared to previous prognostic tasks in lung cancer (CT-based). This extension to neuro-oncology, with fundamentally different tumor behavior and imaging characteristics, highlights our model's generalizability to diverse oncological domains. The model was trained on the UPENN-GBM cohort (n=585) and validated on the BraTS dataset (n=163), with optimization targeting one year OS - a critical benchmark set by previuous works for clinical decision-making in neuro-oncology. In the BraTS validation cohort, \methodname{} achieved an AUC of 0.720 (95\% CI: 0.634–0.800), outperforming the baseline model (AUC: 0.701, 95\% CI: 0.614–0.781) by 2.7\%. Kaplan-Meier analysis stratified patients into distinct prognostic groups ($P$ < 0.001). When compared to GPT-4o, \methodname{} exhibited superior accuracy (0.652; 95\% CI: 0.577–0.730) relative to GPT-4o (0.509; 95\% CI: 0.429–0.583) and more precise risk stratification ($P$ < 0.001 for \methodname{} versus $P$ = 0.365 for GPT-4o), underscoring the necessity of domain-specific adaptation in neuro-oncological AI.

\begin{figure}[!b]
    \centering
    \includegraphics[width=1.0\linewidth]{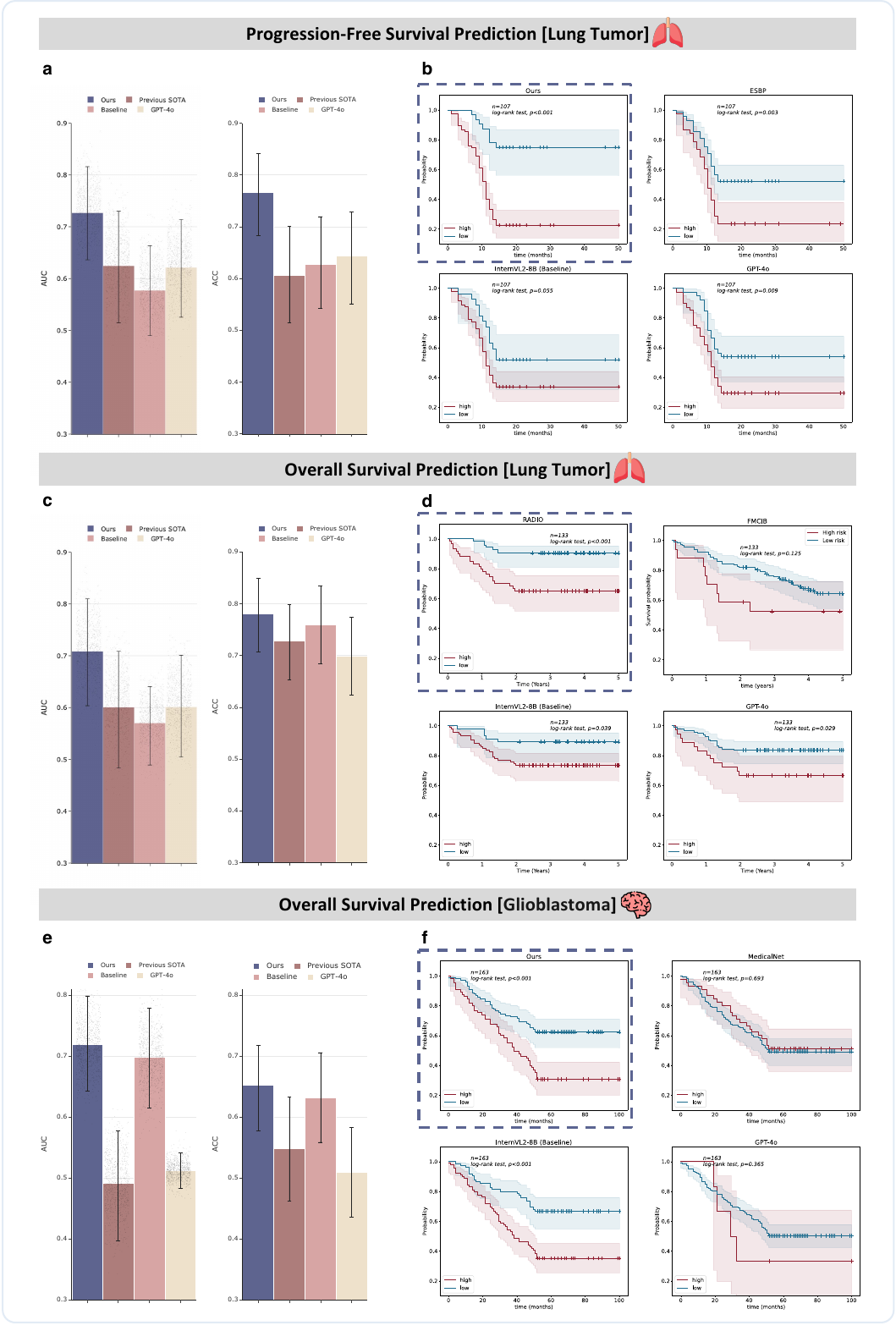}
\end{figure}
\begin{figure}[!t]
    \centering
    \caption{\textbf{a, c, e}, Performance comparison of foundation model-based approaches and baseline methods using AUC and F1 scores. Panel a shows results for 14-month progression-free survival prediction, c for 2-year overall survival prediction in non-small cell lung cancer (NSCLC), and e for 1-year overall survival prediction in glioblastoma. Refer to the Methods section for implementation details. \textbf{d,e,f,}, Kaplan–Meier survival curves stratified by model-predicted risk groups across the three prognostic tasks. Risk thresholds were determined using internal validation sets to ensure fair comparison across methods.}
    \label{fig:page3}
\end{figure}

\subsection*{\methodname{} is A Clinician-Centric AI Assistant for Multi-Modal Interpretation and Trustworthy Support}

\noindent \methodname{} bridges the gap between opaque AI systems and clinically actionable insights by unifying visual-textual evidence, structured reasoning, and reliable uncertainty quantification. This framework ensures clinicians can audit AI outputs efficiently, mitigating risks of misguidance from unverifiable predictions. 

\noindent\textbf{Precise Uncertainty Quantification for Clinically Reliable AI Responses}

\noindent Motivated by the success of semantic entropy in detecting confabulations across various language models and domains \cite{farquhar2024semanticentropy}, we aimed to evaluate its applicability in the medical domain, specifically in Visual Question Answering (VQA) tasks, and to enhance sentence-level uncertainty estimation in medical report captioning. We observed that both Semantic Entropy (SE) and its discrete approximation (Discrete SE) suffer from imprecise estimates when candidate answers are similar—an issue that becomes more pronounced when sampling is limited, as sampling in multimodal large language models (MLLMs) is computationally expensive due to their large parameters. This leads to overconfidence as entropy is underestimated. To address this, we reformulated the semantic entropy calculation to incorporate a more nuanced approach that accounts for response confidence, thereby significantly improving the reliability of uncertainty estimation.

We first evaluated our method across four distinct medical VQA datasets: SLAKE, VQA-RAD, PathVQA, and VQA-Med \cite{ben2019medvqa}. Our enhanced SE approach demonstrated clear improvements over both traditional SE and Discrete SE, with an average performance increase of 9.40\% across all datasets. Notably, on the SLAKE dataset, the area under the curve (AUC) improve 14.1\% from 0.71 to 0.81, and similar improvements were observed on the VQA-RAD dataset (improve 9.1\% from 0.734 to 0.801). These results underscore the robustness and reliability of our method in uncertainty estimation, a critical aspect of high-stakes medical decision-making.

\begin{figure}[t]
    \centering
    \includegraphics[width=1.0\linewidth]{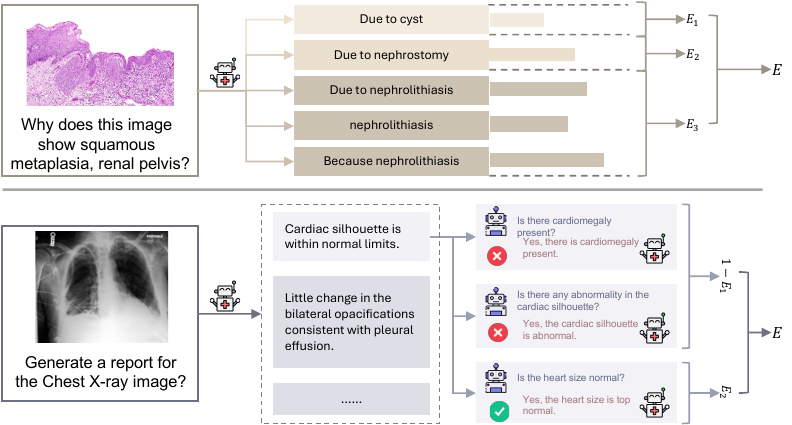}
    \caption{Overview of uncertainty estimation in visual question answering and report generation.}
    \label{fig:Uncertainty_method}
\end{figure}

Having validated the effectiveness of our uncertainty estimation approach on medical VQA tasks, we extended it to the domain of medical report captioning for sentence-level uncertainty estimation, as illustrated in \Cref{fig:Uncertainty_method}. For each sentence generated in a medical report, our system generates multiple questions based on the sentence using a large language model. The consistency between the answers to these questions and the caption is assessed, with semantic entropy calculated to quantify uncertainty for each sentence. We evaluated the reliability of the uncertainty estimates on a subset (n = 200) of the MIMIC-CXR test dataset. As the number of questions increased from 1 to 5, the AUC improved from 0.676 to 0.764, reflecting a 13.02\% enhancement. This substantial improvement demonstrates the effectiveness of combining semantic entropy with test-time scaling for estimating sentence-level uncertainty.

\noindent\textbf{Visual and Textual Interpretation for Clinically trusted AI Responses}

\noindent Our clinical reasoning framework for medical report generation, designed to mirror diagnostic workflows, significantly enhances the interpretability and reliability of AI-generated reports. The five-step process—anatomical structure identification, region localization, region-specific description, holistic analysis, and final report synthesis—integrates patient history, chief complaints, and imaging findings to produce structured, evidence-backed outputs. By explicitly linking clinical context to visual evidence, the model achieves traceable anatomical localization, with region markings demonstrating 95.6\% matchness to reference annotations and a mean Intersection over Union (IoU) of 66.6\%. This visual grounding ensures that early reasoning steps provide spatially precise evidence, a critical advancement over prior methods.

Training on the MIMIC-CXR dataset involved supervised fine-tuning (SFT) with 10,000 samples for long chain-of-thought reasoning, followed by reinforcement learning via Direct Preference Optimization (DPO) on 8,201 high/low-scoring report pairs. Evaluation on a test set annotated with Chest-Imagenome-derived regions revealed a 6.05\% average improvement across clinical metrics (CheXpert F1, Radgraph F1, CheXpert Similarity) compared to non-reasoning baselines. DPO further refined report quality, elevating CheXpert F1 from 0.362 to 0.372.  The model’s reasoning coherence was quantitatively validated using Kendall’s tau-b correlation, which measured alignment between intermediate reasoning paths and final conclusions. Scores of 0.479 (consistency) and 0.474 (completeness; both  $P$ < 0.05) indicate strong logical agreement.

\begin{figure}[!b]
    \centering
    \includegraphics[width=1.0\linewidth]{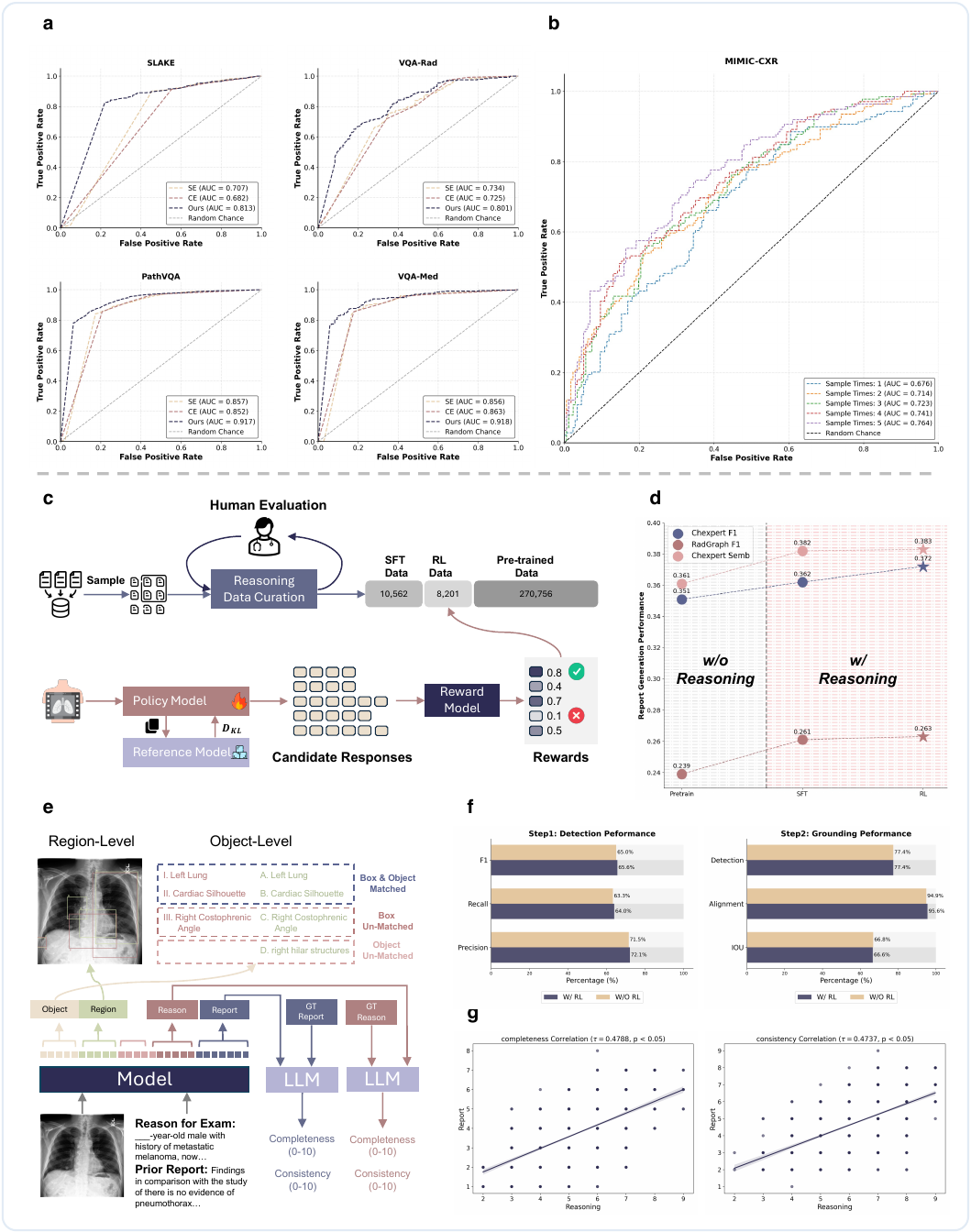}
    \caption{\textbf{a}, Comparison of our method with conventional semantic entropy (SE) and discrete entropy (DE) on four VQA tasks using AUROC to assess error prediction. \textbf{b}, Performance on image captioning tasks, showing our method AUC improves with more generated questions per sentence. \textbf{c}, Overview of curated datasets for supervised fine-tuning (10k samples) and direct preference optimization (8k samples); see \Cref{fig:Prompt_reasoning} and \Cref{fig:Reasoning_data} for details. \textbf{d}, Impact of long chain-of-thought (CoT) reasoning on clinical captioning tasks, evaluated using CheXpert F1, RadGraph F1, and CheXpert Similarity. \textbf{e}, Evaluation protocol for Steps 1, 2, and 4 in CoT: Step 1 and 2 assess region-level, object-level, and alignment performance; Step 4 measures completeness, consistency, and correlation between reasoning and report scores (Kendall Tau). \textbf{f}, DPO maintains localization accuracy in Steps 1 and 2 while improve the reasoning reliability and report accuracy. \textbf{g}, Improved correlation (Kendall Tau) between reasoning and report quality in Step 4 after applying DPO.}
    \label{fig:page4}
\end{figure}

\subsection*{\methodname{} is A Medical-Omni Model Integrating Audio and Regional Functionality for Flexible Human-AI Interaction}
\methodname{} is a groundbreaking Medical-Omni model that seamlessly integrates both audio
and dual regional functionality, revolutionizing human-AI interaction and
expanding its applicability across a broad spectrum of biomedical tasks. By incorporating
audio capabilities, the model facilitates a more intuitive and natural
communication channel between healthcare professionals and the AI system, streamlining
clinical workflows. Meanwhile, the integration of regional inputs allows the model
to perform advanced pixel-level analysis including region recognition and localization
tasks. This dual functionality enables seamless interaction, where users can easily
localize lesions and analyze specific regions of interest, providing a powerful
tool for both healthcare providers and patients.

We rigorously evaluated regional performance of our model across diverse
biomedical tasks, including region recognition, lesion localization, region-based
visual question answering (VQA). We use 1,272 test samples for region recognition and 1,000 test samples for lesion detection test examples from a variety of segmentation and detection datasets to assess region recognition and lesion detection capabilities. For region recognition, the model was tasked with
identifying 141 distinct regions, each provided with a corresponding bounding box.
The F1-score was used as the metric to evaluate its recognition performance. In the
lesion localization task, the model was required to accurately localize lesions
based on their designated names. Intersection over
Union (IoU) was used to assess the localization accuracy. For region VQA, we
tested the model on the Med-GRIT dataset~\cite{huang2024bird}, which contains 30,000 question-answer pairs
across eight different medical imaging modalities. The Med-GRIT dataset includes
four sub-tasks: Referring Object Classification (ROC), Referring Captioning (RC),
Visual Grounding (VG), and Medical Image Analysis (MIA). In line with the BiRD framework,
we employed acc@0.5 to evaluate ROC, SPICE for RC, recall@0.5 for VG, and mBMR (average
of BLEU-2, METEOR and ROUGE-L) for MIA, ensuring comprehensive evaluation across
each task.

\begin{figure}[!b]
    \centering
    \includegraphics[width=1.0\linewidth]{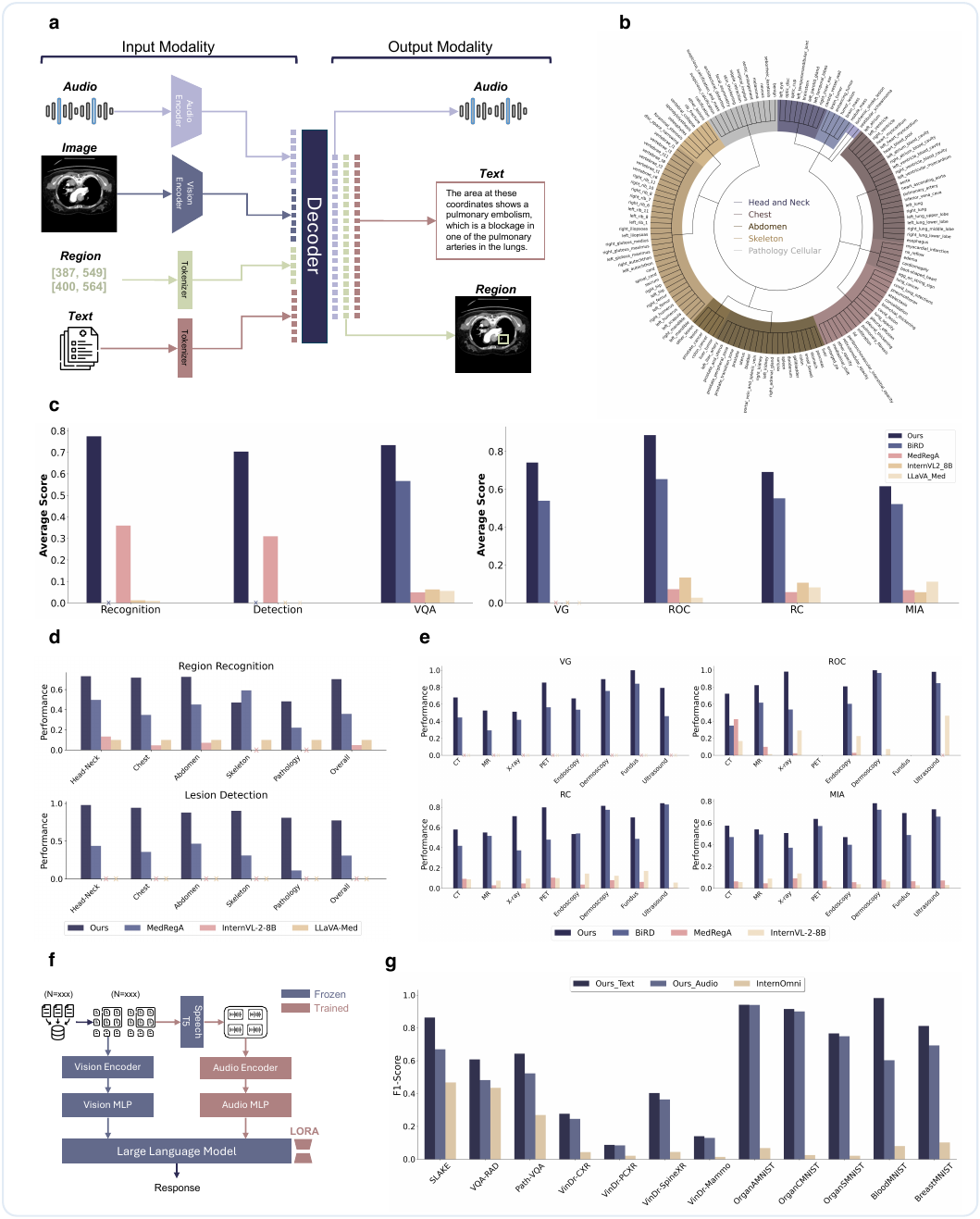}
    \caption{\textbf{a}, \methodname{} is an all-in-one biomedical model capable of processing image, text, audio, and region-level inputs, enabling tasks such as region recognition, region-based VQA, and lesion localization. \textbf{b}, List of anatomical regions recognizable by \methodname{}, which supports localization of 141 lesion types (see \Cref{fig:page5}b for details). \textbf{c}, Performance on region-related tasks: F1-score for region recognition, IoU for lesion localization, and task-specific metrics for Region VQA—acc@0.5 (ROC), SPICE (RC), recall@0.5 (VG), and mBMR (BLEU-2, METEOR, ROUGE-L) for MIA, following the BiRD convention. \textbf{d}, Comparative results on region recognition and lesion localization across six anatomical structures. \textbf{e}, Performance comparison on Region VQA tasks across eight imaging modalities. \textbf{f}, Overview of audio data curation and fine-tuning: 514,668 pretraining instances with audio input were used to update the audio encoder, audio MLP, and LLM (via LoRA), with other components frozen. \textbf{g}, Comparison of \methodname{} with audio vs. text inputs and with leading audio models on single-label diagnosis, multi-label diagnosis, and medical VQA tasks.}
    \label{fig:page5}
\end{figure}

% Region Part
The overall performance across the three primary tasks is illustrated in \Cref{fig:page5}.  While previous models, such as MedRegA~\cite{wang2024medrega} and BiRD~\cite{huang2024bird}, have introduced anatomical localization into MLLMs,
their performance falls short when compared to our \methodname{}. For region recognition,
\methodname{} achieved an F1-score of 0.774, surpassing SOTA approach MedRegA F1-score
0.359, lowering our method by 53.6\% , demonstrating its superior ability to accurately
identify and analyze regions. In anatomical localization, \methodname{} achieved an IoU of
0.703, outshining MedRegA 0.309 by a large margin, underscoring its precision in
accurately pinpointing lesions (See \Cref{tab:region_rec_lesion_det} for detail). In the region VQA task, \methodname{} achieved an average
score of 0.733 across four sub-tasks, outperforming the previous state-of-the-art
model, BiRD, by 29.4\%, with improvements ranging from 18.0\% to 40.6\% across
the individual sub-tasks (See \Cref{tab:MedGRIT} for detail). These results highlight the advanced capabilities of \methodname{}
in pixel-level analysis across diverse biomedical tasks. Beyond general
performance, we also assessed the model’s ability to handle different anatomical
structures. \methodname{} consistently outperformed MedRegA across all structures
except for the skeleton, where it was slightly behind. For lesion localization, \methodname{}
excelled in all anatomical structures, surpassing MedRegA across the board. In region
VQA, the model again demonstrated superior performance, outperforming BiRD across
all structures. Furthermore, when assessed across different modalities, \methodname{}
maintained its superiority, outpacing BiRD in all modalities except for Endoscopy
in the Referring Captioning (RC) task, where it performed slightly below BiRD (lower
by 1\%). Notably, in CT, X-ray, and Fundus imaging, \methodname{} led BiRD by margins
ranging from 34.5\% to 58.1\% averaged on four tasks. These results underscore the
robustness, adaptability, and generalizability of \methodname{} in pixel-level analysis
and biomedical task performance.

% Audio part
Incorporating audio capabilities further amplifies the model's versatility. We fine-tuned
the model with a subset of the pre-trained dataset, converting text into audio
using the Speechh5 model~\cite{ao2021speecht5}, which includes a diverse array of audio styles. Our model
with audio input surpasses the baseline model for a large margin, achieving an
average improvement of 0.399 (from 0.133 to 0.532) across all benchmarks
compared to the baseline method (Intern-Omni)~\cite{chen2024internomni} (See \Cref{tab:audio} for detail). Although audio input slightly lagged
behind text input across most benchmarks, it performed equally well in more
complex multi-label diagnostic tasks, only lowering by 0.044 (from 0.250 to
0.206) avereaged on four multi-label diagnosis tasks (VinDr-CXR, VinDr-PCXR,
VinDr-SpineXR, VinDr-Mammo), demonstrating the reliability and robustness of audio
input. These results highlight the potential of combining audio and regional
inputs to offer a more flexible, robust, and interactive model for medical
applications, enhancing the model’s overall utility in human-AI collaboration.

\subsection*{\methodname{} Unleashing Robust Generalizability Across Diverse Tasks and Ensuring Reliable AI with Guardrail Capabilities}

The external validation on in-house data with different distribution with the training
data on diverse tasks the efficient application of a single, pre-trained model
to a wide array of tasks, without the need for extensive retraining or data annotation
for each new distribution. \methodname{}, pre-trained on a large-scale medical dataset incorporates over 114 open-source datasets, exhibits exceptional
generalizability across various tasks without the need for further fine-tuning. We
evaluated \methodname{}’s generalizability on in-house data collected from Sun Yat-sen Memorial Hospital, Sun Yat-sen University, which contains well-annotated masks for 11 clinically significant diseases. We
assessed its performance on four primary tasks: Multiple-Choice, Diagnosis, Localization,
and Visual Question Answering (VQA). For Multiple-Choice tasks, we use accuracy
to evaluate performance. For Diagnosis, we use the F1 score, and for Localization,
we assess performance using IoU. In VQA, we employ the LLM-as-Judge approach, evaluating
the generated answers across three aspects: accuracy, consistency, and clinical
relevance as the NLG metric alone does not fully capture semantic similarity
with the reference.

For Multiple-Choice tasks, \methodname{} achieved an accuracy of 94.0\%, outperforming
MedRegA's 80.6\% by 16.6\%. Notably, the baseline model InternVL2-8B showed significantly
better performance compared to large-scale medical models such as LLaVA-Med,
RadFM, Medflamingo, and MedDr, indicating that existing medical MLLMs struggle with
generalizing to new distributions, even in simpler tasks like multiple-choice. In
the Diagnosis and Localization tasks, \methodname{} also performed exceptionally, achieving
an F1 score of 0.768 and an IoU of 0.431, surpassing MedRegA's 0.189 and 0.221
by a remarkable margin. For VQA task (See \Cref{tab:external_vqa} for detail), \methodname{} achieved average scores of 8.61
across three evaluation aspects, outperforming previous SOTA MedDr by 7.9\%. Notably,
when evaluating accuracy, consistency, and clinical relevance—the key metrics
for assessing the correctness of generated answers—\methodname{} demonstrated a
marked advantage, improving 7.2\%, 6.9\%, and 8.6\% compared to previous SOTA, indicating
its reliability and robustness in VQA tasks.

We further assessed \methodname{}'s performance in Region Recognition and Lesion Localization
across 11 clinically significant diseases, using both internal and external
validation datasets (See \Cref{tab:external_regionrec_lesiondet} for detail). In Region Recognition, \methodname{}'s performance on the
external dataset was slightly lower by 0.12 on average compared to the internal
dataset. The model’s performance was less robust in kidney and prostatic cancers.
However, in certain diseases, such as cerebral hemorrhage and pancreatic cancer,
the model outperformed the internal dataset, showcasing its strong
generalizability. Similarly, for Lesion Localization, \methodname{} showed a slight decrease
of 0.19 on the external dataset, but excelled in diseases like cerebral
hemorrhage and pancreatic cancer, surpassing internal dataset results. Notably, in
the challenging task of detecting Pulmonary Embolism, \methodname{} showed a drop of
0.44 on the external dataset. For liver cancer, the model achieved
an improvement of 9.2\% in IoU for lesion localization. These results highlight the
robust generalizability of \methodname{} across diverse clinical tasks and diseases,
demonstrating its ability to maintain reliable and consistent performance on
unseen data. This ensures its applicability in real-world clinical settings,
making it a highly effective and adaptable tool for medical applications.

In addition to its remarkable generalizability, we also assessed the guardrail capabilities
of \methodname{}—an essential feature that ensures the model’s reliability and safety
in clinical environments. Guardrails are vital for preventing the model from
making incorrect or unsafe predictions, particularly in biomedical tasks. For
instance, when given image-specific instructions such as “Detect the right kidney
from the image,” if no kidney is present, the model is expected to respond
appropriately, saying, “Sorry, I cannot find the right kidney in the image.” We refer
to such cases as "mismatch scenarios". \methodname{} demonstrated exceptional
guardrail performance, successfully detecting 83.02\% of mismatch scenarios, while
previous models failed to identify any. This impressive result underscores the
model’s capacity to detect potential mismatches and ensure reliable, safe predictions,
highlighting its suitability for real-world clinical applications.

\begin{figure}[!b]
    \centering
    \includegraphics[width=1.0\linewidth]{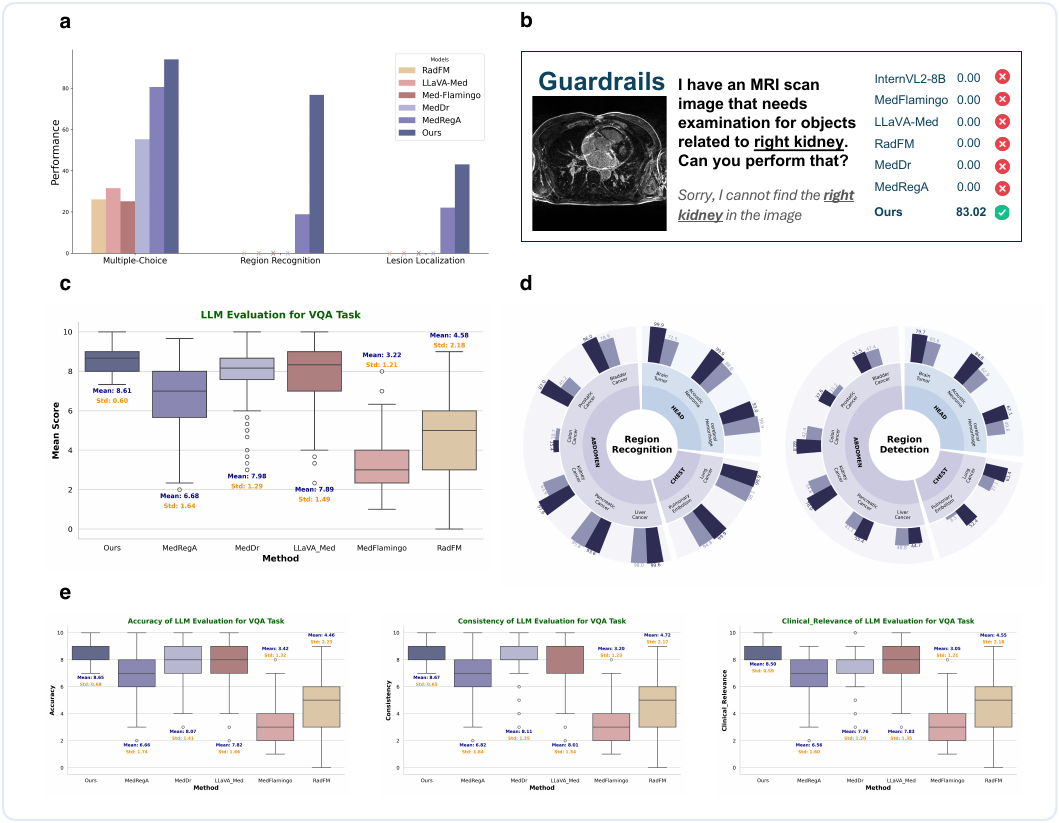}
    \caption{\textbf{a}, Evaluation on the external
    validation data across diverse biomedical tasks (Multiple-choice, Region Recognition
    and Lesion Localization). Accuracy is used as the evaluation metric for multiple-choice,
    F1-score is used for region recognition, and Intersection over Union (IoU) is
    used for lesion localization. \textbf{b}, Evaluation of the guardrail
    capability of \methodname{} and other leading models, with accuracy in detecting unanswerable
    questions from the provided images used as the evaluation metric. \textbf{c},
    Results for 11 clinically significant diseases from diverse anatomical
    regions, evaluated on internal and external datasets. The results
    demonstrate the generalization performance of \methodname{} on various diseases. \textbf{d},
    Comparison of \methodname{} with leading models on Region VQA tasks. LLM is used
    to evaluate the generated free-text from three aspects: accuracy,
    consistency, and clinical relevance.}
    \label{fig:page6}
\end{figure}
\section*{Discussion}

In this study, we present \methodname{}, a new Medical Multimodal Large Language
model expanding the capabilities of Generalist Medical AI (GMAI) to previously
underexplored yet clinically significant decision-making tasks and enabling
seamless collaboration with flexible interaction modes and trustworthy responses.
Through extensive evaluation on 38 benchmarks covering 18 medical modalities, we
show that \methodname{} achieves superior performance over existing GMAI models,
without requiring any further fine-tuning. While previous GMAI models have focused
on relatively simple biomedical tasks, such as image classification, visual question
answering and captioning, with intended applications for diagnosis and lesion
detection, \methodname{} firstly extends the capabilities of GMAI to more
complex clinical decision-making tasks by using its ability to synthesize
information from multimodal data sources, such as radiography, patient
demographics and medical history. The superior performance on progression-free survival
prediction and overall survival prediction tasks demonstrated that the multimodal
fusion capabilities of \methodname{} can be effectively leveraged to improve the
clinical decision-making process. In addition, we conducted sufficient external
validation across diverse modalities, diseases, and tasks to evaluate the
generalizability and robustness of \methodname{} considering the heterogeneity of
data distribution across different sources. The results show that \methodname{} can
generalize well to unseen data and tasks, demonstrating its potential for real-world
clinical applications. However, we observed that the external validation
performance on detecting Pulmonary Embolism showed a significant performance
drop compared to the internal validation results. This observation suggests that
the adaptation to small target detection is still a challenging task for GMAI models
and requires further investigation to attain comprehensive generalizibility for GMAI
models. Importantly, \methodname{} holds presents impressive guardrail
capabilities for preventing the model from making incorrect or unsafe
predictions, ensureing the model’s reliability and safety in clinical environments.

To foster practical clinical applications with human-AI collaboration, a major
challenge is to equip the model with flexible interaction modes and
interpretable, actionable and easily verifiable responses. Our \methodname{} introduce
a mechanism that combines the model’s responses with sentence-level uncertainty,
detailed diagnostic reasoning, and grounding evidence, providing an effective
solution for the model to communicate with human experts in a more interpretable
and trustworthy manner. We observed the reliability of estimated uncertainty improved
with the scaling of the sampled question-answer pairs, which suggests that the
test-time scaling of the model can be an effective strategy to harness the
model’s uncertainty estimation capabilities. Furthermore, we demonstrated that
the response associated with detailed diagnostic reasoning and grounding
evidence not only provides a more interpretable and actionable response but also
improve the performance on the challenging captioning tasks. In addition, fine-tuning
on large-scale image-level and pixel-level annotations and further training on
audio data provided the model with the ability to support a variety of input-output
formats such as images, text, audio, and region-based inputs—offering enhanced adaptability
and interaction. The evaluation on diverse regional-related tasks showed that
\methodname{} can effectively handle region-based inputs and detect more than 200
regions across all anatomical structures. Our model with audio input also
attained much superior performance than baseline model and achieved competitive
performance with text input. Both of the results suggest that \methodname{}
demonstrate its potential for providing flexible interaction modes for human-AI
collaboration.

Despite promising results in clinical decision-making tasks, such as progression-free
survival (PFS) and overall survival (OS) prediction, the model's performance was
evaluated on relatively small cohorts—49 and 58 patients for PFS from two
centers, 133 patients for lung cancer overall survival prediction from one center, and 163 patients for glioma tumor overall survival prediction from one center. Thus, further validation in
larger, multi-center cohorts is necessary before clinical implementation,
particularly for high-stakes applications, which will require regulatory approval
through prospective trials involving diverse patient populations. Additionally, while
the model performed well on a variety of external validation tasks, it showed a
significant drop in performance for small target detection tasks, such as pulmonary
embolism and kidney and prostate cancer recognition. This highlights the challenge
of adapting generalist medical AI to small target detection, and future research
should address issues related to data diversity, resolution, and detection algorithms.
Moreover, the model currently lacks the capability to process video inputs or 3D
medical images, which are common in medical imaging. Expanding the model to handle
these formats requires collecting and annotating large-scale video and 3D datasets,
and future work should focus on developing strategies to generate and leverage
such datasets to enhance the model’s ability to process 3D medical images.

\section*{Conclusion}
\methodname{} introduces a clinician-centric paradigm for Generalist Medical AI (GMAI) that addresses critical limitations in multi-modal explainability, prognostic accuracy, and human-AI collaboration. By combining visual lesion localization with text-based diagnostic reasoning, alongside a novel Reliability Indexing mechanism that quantifies prediction uncertainty, \methodname{} significantly reduces AI-induced errors by 15.2\%, allowing clinicians to confidently validate high-stakes outputs. Trained on a diverse dataset of 7 million multi-modal medical data points, \methodname{} excels in detecting subtle longitudinal tumor changes and synthesizing multi-visit data for prognostic decision-making. It achieves a 26.9\% average improvement in AUC over previous leading models for overall survival and progression-free survival prediction, encompassing both high-incidence cancers (e.g., lung adenocarcinoma) and high-mortality diseases (e.g., glioblastoma). Its robust generalizability across 141 anatomical regions and 40 modalities, validated through extensive external trials, positions \methodname{} as the first GMAI with scalable, clinician-ready utility. This system not only advances precision oncology but also establishes a new benchmark for AI tools that enhance diagnostic autonomy while preserving clinical expertise, driving the integration of trustworthy AI into real-world care pathways.
\section*{Methodology}

\subsection*{Dataset Curation for Pretraining}
% Dataset Statistics and Sources
For the pretraining of the Generalists Medical AI \methodname{}, we utilized a diverse
collection of publicly available medical datasets to create a comprehensive,
large-scale dataset tailored for image-level tasks, region-level tasks, and
clinical applications. The image-level dataset comprises 5,468,115 pairs of images and
corresponding questions. This dataset is designed to support tasks such as medical diagnosis, visual question answering, and
image captioning. The pixel-level dataset includes 1,686,677 pairs of images and
questions, each annotated with regional information. This pixel-level dataset encompasses 141 distinct regions of interest. The inclusion of these medical tokens allows the model to process
a wide range of biomedical instructions and interpret long-context information,
thereby enhancing its capacity for nuanced understanding across different
medical contexts.

\begin{figure}[t]
    \centering
    \includegraphics[width=1.0\linewidth]{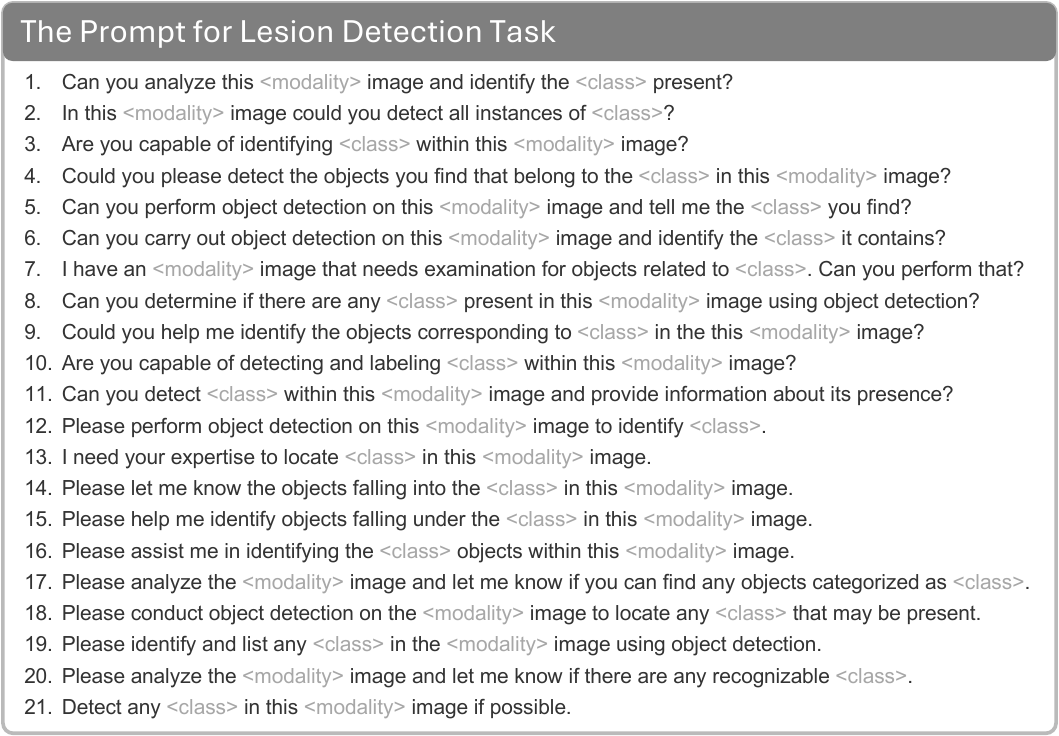}
    \caption{Illustration of the prompt design used for lesion detection task.}
    \label{fig:Prompt_Detect}
\end{figure}

\begin{figure}[t]
    \centering
    \includegraphics[width=1.0\linewidth]{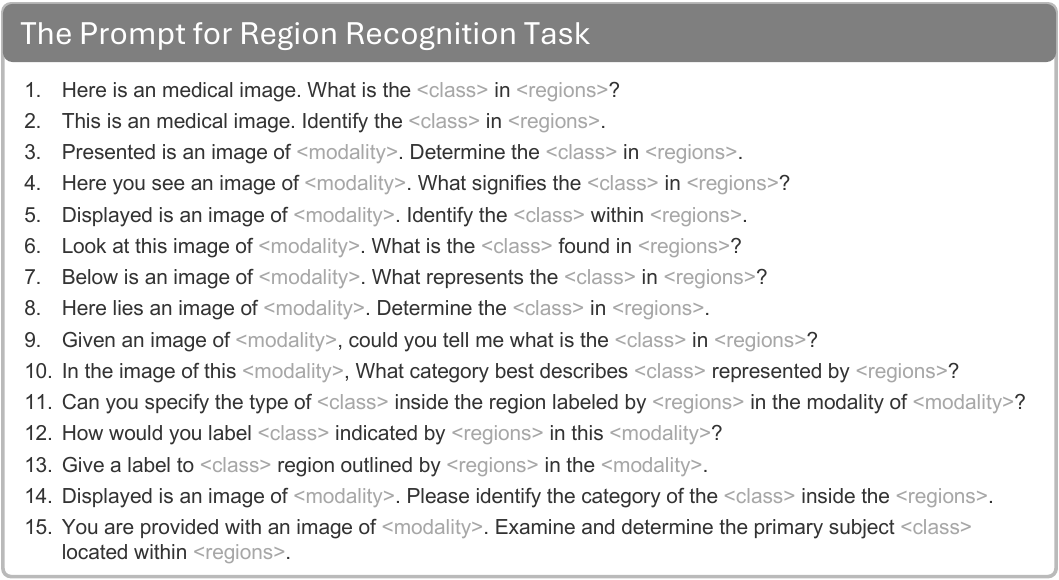}
    \caption{Illustration of the prompt design used for region recognition task.}
    \label{fig:Prompt_RegionRec}
\end{figure}

% Prompt Design for Different Tasks
To support the pretraining of \methodname{} across various medical tasks, we developed
a set of tailored prompts designed for different medical applications. Previous
research has shown that diverse prompts during pretraining can significantly
improve the model's ability to follow instructions and enhance its performance
on downstream tasks. The prompts for each task are illustrated in \Cref{fig:Prompt_Detect} and \Cref{fig:Prompt_RegionRec}.
For each task, we begin by providing a template and then specify the task-specific
instructions. To further diversify the prompt styles, we employ large language models
(e.g., GPT-4) to generate a wide range of variations in the task instructions.
For example, for the lesion detection task, we start with the template: ``Can you
analyze this <modality> image and identify the <class> present?", where <modality>
represents the imaging modality (e.g., X-ray, MRI) and <class> denotes the type of
lesion (e.g., tumor, fracture). We then use GPT-4 to generate diverse additional,
varied instructions, such as: ``In this <modality> image, could you detect all instances
of <class>?" and ``Please locate and identify all visible <class> in the <modality>
scan." These instructions provide different ways of asking the same task,
ensuring that the model is exposed to a wide range of phrasing and perspectives.
By incorporating this diversity of prompts, we aim to improve the model's
ability to understand and follow task-specific instructions, thereby making it a
versatile and user-friendly tool for a broad spectrum of medical applications.

\subsection*{Model design and pretraining}
The training of \methodname{} is conducted using a multi-modal transformer
architecture that combines vision and language modalities. The model architecture
is based on InternVL-2, a state-of-the-art multi-modal transformer model that
has been pre-trained on a large-scale dataset of image-text pairs. As demonstrated
in previous studies, supervised fine-tuning on pre-trained models for domain-specific
applications achieves superior performance compared to training from scratch. To
adapt the model to the medical domain, we fine-tune the InternVL-2 model on our
large-scale medical dataset using a supervised learning approach for one epoch.

\subsection*{Benchmark datasets}
We evaluated \methodname{} across 23 benchmarks, spanning a wide range of medical
tasks including single-label diagnosis, multi-label diagnosis, multiple-choice questions,
visual question answering (VQA), and image captioning. These benchmarks were
based on various publicly available biomedical datasets. For single-label diagnosis,
we used nine datasets, covering 34,821 samples for organ recognition (OrganAMNIST
(n=17,778), OrganCMNIST (n=8,216), OrganSMNIST (n=8,827)) and 13,193 samples for
lesion recognition (PathMNIST (n=7,179), OCTMNIST (n=999), PneumoniaMNIST (n=623),
BreastMNIST (n=155), PAD-UFES-20 (n=1,044), MURA (n=3,193)). These datasets focus
on identifying specific organs or lesions from medical images, testing the model's
capacity to perform classification tasks. In multi-label diagnosis, we assessed
the model's performance on six lesion classification datasets. These included 2,077
samples for spinal lesions (VinDr-SpineXR), 8,387 samples for thoracic diseases
(VinDr-CXR (n=3,000), VinDr-PCXR (n=1,387), VinDr-Mammo (n=4,000)), and 3,894 samples
for ocular lesions (RFMiD (n=640), BRSE (n=3,254)). Multi-label classification tasks
require the model to identify multiple lesions within a single image, providing
a more complex challenge for lesion recognition. The visual question answering benchmarks
included two commonly used radiology datasets, which together cover 1,512
samples across five anatomical regions (VQA-Rad (n=451), SLAKE (n=1,061)), and
an additional 6,761 samples from a pathology dataset (PathVQA), which combines anatomical
and tissue-specific details. These tasks test the model's ability to understand
and respond to medical questions about images, ranging from simple anatomical recognition
to more complex tissue-specific inquiries. To further evaluate the model’s
generalizability across clinical tasks and modalities, we also tested it on two comprehensive
multiple-choice benchmarks: OmnimedVQA and GMAI-MMbench. OmnimedVQA includes 72,683
samples sourced from 42 datasets, representing 8 modalities, while GMAI-MMbench encompasses
21,281 samples derived from 284 datasets. The latter covers a broad spectrum of
38 medical image modalities, 18 clinical tasks, 18 medical departments, and 4 levels
of perceptual granularity. For image captioning, we evaluated \methodname{} on
two widely used chest X-ray (CXR) datasets: IU-Xray (n=1,180) and Medical Information
Mart for Intensive Care III-CXR (MIMIC-CXR) (n=4,710). These datasets test the
model’s ability to generate descriptive captions for medical images, a key task in
enhancing automated medical reporting.

\subsection*{Implementation for porgnostic decision making}

\noindent
\textbf{Progression-free survival prediction for non-small cell lung cancer.} For this task, we utilized multi-center datasets collected from four institutions: the Second Affiliated Hospital of Zhejiang University School of Medicine (ZJ, n=238), the Affiliated Hospital of Jiaxing University (JX, n=58), Southern Medical University (NF, n=85), and Hunan Cancer Hospital (HN, n=49). The ZJ and NF cohorts were combined as the primary dataset, with an 80\%:20\% split for training and internal validation. The JX and HN cohorts were used exclusively for external validation. Each patient's data comprised baseline lung CT scans (prior to treatment), smoking history, TNM stage, metastatic status, mutation status, and details of tyrosine kinase inhibitor (TKI) treatment. For some patients, longitudinal follow-up data were available, including additional lung CT scans taken at 1 month and 3 months in treatment. Progression-free survival (PFS) was defined as the duration from the date of initial drug administration to the time of either disease progression or death, whichever occurred first. PFS served as the primary outcome measure for assessing additional survival benefit, with the prognostic performance of ESPS~\cite{deng2022deep_ESBP} evaluated accordingly. Patients who remained alive without recorded progression were censored at the date of their last follow-up. Based on follow-up assessments, the median PFS for patients receiving EGFR-TKI treatment was determined to be 14 months. Patients were categorized into low-risk (good responders, PFS longer than median) and high-risk groups (poor responders, PFS shorter than median). Patients in the low-risk group were considered to have gained additional survival benefit, whereas those in the high-risk group were not considered to have gained additional survival benefit from EGFR-TKI therapy. For implementation details, our proposed method and baseline methods (InternVL2-8B and GPT-4o) leveraged these multi-visit CT scans, when available, to longitudinally assess disease progression. For the ESPS model specifically, we adhered to its standard protocol, initially cropping tumor regions from CT images before performing continuous fine-tuning on the ZJ and NF training datasets.

\noindent \textbf{Overall survival prediction for non-small cell lung cancer.} To evaluate overall survival (OS) in patients with non-small cell lung cancer (NSCLC), we utilized two publicly available datasets: LUNG1 and RADIO. The LUNG1 dataset comprises 422 patients with stage I–IIIB NSCLC who underwent radiation therapy at MAASTRO Clinic (Maastricht, the Netherlands). For this study, we selected 420 patients with annotated gross tumor volumes and survival data right-censored at 2 years. The selected cohort was used exclusively for tuning the models. The RADIO dataset includes 211 NSCLC patients (stage I–IV) referred for surgical treatment between 2008 and 2012. Following the convention of FMCIB, a subset of 144 patients includes tumor segmentations reviewed by two thoracic radiologists, and molecular profiling data (e.g., EGFR, KRAS, ALK mutations, gene expression arrays, RNA sequencing) are available for a majority of the cohort. Finally, we used a subset of 133 patients with annotated tumor volumes and survival data (right-censored at 2 years) as an independent test set. To assess prognostic performance, we classified patients into two risk categories based on their overall survival duration: those who survived beyond two years were designated as low-risk, while those who survived less than two years were considered high-risk. or model implementation, our proposed method and two baseline methods (InternVL2-8B and GPT-4o) utilized patient-level CT slices containing tumor regions as input, combined with textual clinical information. The models estimated the probability of a patient surviving longer than two years by averaging the predictions across all CT slices. For FMCIB, we followed the original implementation by extracting features from the foundation model and training a linear classifier using the LUNG1 cohort.

\noindent\textbf{Overall survival prediction for glioblastoma.} To assess overall survival in patients diagnosed with glioblastoma, we employed two publicly available datasets: UPEEN-GBM (n=585) and TCGA-GBM/BraTS (n=163). The UPEEN-GBM dataset was used for training and internal validation, following an 80\%:20\% split. The TCGA-GBM/BraTS cohort served as an independent external validation set. One-year overall survival (OS) was selected as the primary endpoint for all analyses. Both datasets provide four standard MRI modalities per patient: T1-weighted (T1), T1-weighted post-contrast (T1CE), T2-weighted (T2), and fluid-attenuated inversion recovery (FLAIR) images. For consistency and fair comparison across methods, we selected FLAIR as the primary imaging modality for training and evaluation. The model implementation followed the same framework as the NSCLC OS prediction task. Specifically, our approach and baseline methods (InternVL2-8B and GPT-4o) processed the FLAIR slices containing tumor regions, supplemented with available clinical features, to predict whether a patient's survival exceeded one year. Slice-level predictions were aggregated to produce patient-level survival probabilities. Additionally, we employed MedicalNet as a task-specific model; this foundation model was pre-trained on a diverse set of medical imaging modalities and tasks, providing strong generalizability for downstream survival prediction.
% Medical-Omni post-training

\subsection*{Formulation for robust uncertainty estimation}
\noindent
\textbf{Formulation of Semantic Entropy.} The estimation for semantic entropy comprises
three steps: (1) Sampling output sequences of tokens from the predictive
distribution of the model given the input image $\vx$ and query
$\textbf{\text{q}}$. (2) Clustering generated sequences by their semantic similarity
to obtain the cluster assignments for each sample. (3) Estimating semantic
entropy by summing probabilities of the sequences in the same cluster. Specifically,
given the input image $\vx$ and query $\textbf{\text{q}}$ as input
to \methodname{}, we sample $M$ output sequences of tokens, $\{\vs^{(1)}
, \vs^{(2)}, \ldots, \vs^{(M)}\}$, and their
corresponding token probabilities, $\{P( \vs^{(1)}| \vx, \textbf{\text{q}})\}), P (\vs^{(2)}| \vx, \textbf{
 \text{q}}), \ldots, P(\textbf{\text{s} }^{(M)}| \vx, \textbf{\text{q}
})\}$. We sample all generations from \methodname{} with a temperature of 1.0 and
a top-p sampling of 0.9. Next, we cluster the generated sequences into $C$ groups
that share the same semantic meaning. A semantic equivalence function
$E(\cdot, \cdot)$ is used to determine the semantic equivalence (entailment)
between two sequences as follows:
\begin{equation}
    E(\vs^{(i)}, \vs^{(j)}) =
    \begin{cases}
        1, & \text{if }\vs^{(i)}\text{ and }\vs^{(j)}\text{ are semantically equivalent}, \\
        0, & \text{otherwise}.
    \end{cases}
\end{equation}
By calculating the semantic equivalence between all pairs of sequences, we can
distribute the samples into $C$ clusters. Each semantic equivalence class $c \in C $ contains sequences that are semantically equivalent,
such that for any two sequences $\vs$ and $\vs'$ in
the same cluster $c$, the equivalence $E(\vs, \vs') =
1$ holds. The semantic equivalence function leveraged the idea of bidirectional
entailment mentioned in \cite{farquhar2024semanticentropy}. That is, a sequence $\vs$ is considered semantically equivalent
to another sequence $\vs'$ if and only if $\vs$ entails
$\vs'$ and vice versa. In this study, we used the pre-trained DeBERTa-Large-MNLI
model to calculate the entailment between two sequences. Having determined the
semantic clusters $C$, we calculate the likelihood that a sequence generated by \methodname{}
belongs to each cluster by computing the sum of the probabilities of the
sequences as follows:
\begin{equation}
    \centering
    P(c| \vx, \textbf{\text{q}}) = \sum_{\vs\in
    c}P(\vs| \vx, \textbf{\text{q}})
    = \sum_{\vs\in c}\prod_{t=1}^{T}P(s_{t}| \vx, \textbf{\text{q}}, \vs_{1:t-1}),
\end{equation}
where $P(c| \vx, \textbf{\text{q}})$ is the probability
of cluster $c$ given the input image $\vx$ and query
$\textbf{\text{q}}$, and
$P(\vs| \vx, \textbf{\text{q}})$ is the probability
of sequence $\vs$ given the input image $\vx$ and
query $\textbf{\text{q}}$. The semantic entropy $H$ is then calculated as:
\begin{equation}
    \centering
    H = - \sum_{c} P(c| \vx, \textbf
    {\text{q}}) \log P(c| \vx, \textbf{\text{q}}),
\end{equation}
where $P(c| \vx, \textbf{\text{q}})$ is the
probability of cluster $c$. The semantic entropy $H$ is estimated
by using the Rao-Blackwellized Monte Carlo integration over the clusters as follows:
\begin{equation}
    \centering
    H = -\sum_{i=1}^{|C|}P(C_{i}| \vx, \textbf{\text{q}}) \log P(C_{i}| \vx, \textbf{\text{q}}),
\end{equation}
where
$P(C_{i}| \vx, \textbf{\text{q}})=\frac{P(C_{i}|
\vx, \textbf{\text{q}})}{\sum_{j=1}^{|C|}P(C_{j}|
\vx, \textbf{\text{q}})}$
and
$\sum_{i=1}^{|C|}P(C_{i}| \vx, \textbf{\text{q}})=
1$. However, when all samples belong to the same group or class, the benefit of Rao-Blackwellization
diminishes because the method relies on the idea of conditioning on sufficient statistics
(or sufficient information) to reduce variance in the estimate, i.e.,
$P(C_{i}| \vx, \textbf{\text{q}}) = 1$ for some
$i$. In this case, the semantic entropy is highly biased and underestimates the
true uncertainty without using the likelihood of the generated sequences. To address
this issue, we propose a novel method to estimate the semantic entropy by
considering the likelihood of the generated sequences when all samples belong to
the same cluster. We first considering the entropy in each cluster as follows:
\begin{equation}
    \centering
    H_{c}= -P(c| \vx, \textbf{\text{q}
    }) \log P(c| \vx, \textbf{\text{q}}) = - \sum
    _{\vs \in c}\frac{P(\vs|\vx,\textbf{\text{q}})}{
    \sum_{\textbf{\text{s}'} \in c}P(\vs'|\vx,\textbf{\text{q}})}
    \cdot \log\left( \frac{P(\vs|\vx,\textbf{\text{q}})}{\sum_{\textbf{\text{s}'}
    \in c}P(\vs'|\vx,\textbf{\text{q}})}
    \right)
    \label{eq:equation5}
\end{equation}
The semantic entropy is then calculated as:
\begin{equation}
    \centering
    H = \sum_{i=1}^{|C|}P(C_{i}| \vx, \textbf{\text{q}
    }) H_{C_{i}}.
\end{equation}
This formulation ensures that the semantic entropy is accurately estimated when the
sampling times $M$ is small or when all samples belong to the same cluster,
providing a more robust measurement of uncertainty in the model's predictions. To
verify the potential limitations of the Rao-Blackwellized Monte Carlo integration
used in the original formulation, we conducted ablation studies on the four VQA
benchmarks with sampling times $M=5, 10, 15, 20, 30$. The results showed that when
the sampling times $M$ is small, the Rao-Blackwellized Monte Carlo integration
underestimates the semantic entropy, leading to lower AUC scores. As the
sampling times $M$ increases, the AUC scores improve, and perform on par with
the proposed method. However, the proposed method consistently outperforms the
Rao-Blackwellized Monte Carlo integration across all sampling times, demonstrating
its robustness and effectiveness in estimating semantic entropy. In addition, we
extracted the $(\vx, \textbf{\text{q}})$ pairs from the VQA benchmarks
that have one cluster and compared the semantic entropy estimated by the Rao-Blackwellized
Monte Carlo integration and the proposed method. The results showed that the
proposed method achieved a higher AUC score than the Rao-Blackwellized Monte
Carlo integration, indicating that the proposed method is more accurate in estimating
the semantic entropy when all samples belong to the same cluster. These results
demonstrate the effectiveness of the proposed method in estimating semantic
entropy, particularly when the samples are highly clustered or when the sampling
times are limited. The results showed that the proposed method estimated more
accurate semantic entropy than the Rao-Blackwellized Monte Carlo integration, particularly
when the samples are highly clustered.

% and then calculate the semantic entropy as follows:
% \begin{equation}
%     \centering
%     H = -\sum_{i=1}^{|C|}P(c_{i}| \vx, \textbf{\text{q}
%     }) \log P(c_{i}| \vx, \textbf{\text{q}}) + \sum_{i=1}^{|C|}P(c_{i}| \vx, \textbf{\text{q}}) \log P(\vs| \vx, \textbf{\text{q}}),
% \end{equation}
% where $P(\vs| \vx, \textbf{\text{q}})$ is the likelihood
% of the generated sequences. This formulation ensures that the semantic entropy
% is accurately estimated even when all samples belong to the same cluster, providing
% a more robust measure of uncertainty in the model's predictions.

% Specifically, we calculate the semantic entropy as follows:
% \begin{equation}
%     \centering
%     H = -\sum_{i=1}^{|C|}P(c_{i}| \vx, \textbf{\text{q}
%     }) \log P(c_{i}| \vx, \textbf{\text{q}}) + \sum_{i=1}^{|C|}P(c_{i}| \vx, \textbf{\text{q}}) \log P(\vs| \vx, \textbf{\text{q}}),
% \end{equation}
% where $P(\vs| \vx, \textbf{\text{q}})$ is the likelihood
% of the generated sequences. This formulation ensures that the semantic entropy
% is accurately estimated even when all samples belong to the same cluster, providing
% a more robust measure of uncertainty in the model's predictions.

% Our improved formulation and implementation of semantic entropy estimation
\noindent
\textbf{Sentence-level uncertainty estimation for medical report generation.}
We extend this methodology to sentence-level uncertainty estimation in medical
report generation. Robust and accurate uncertainty estimation in this context is
essential for ensuring the reliability of generated reports and facilitating human-AI
collaboration. In contrast to short-length generation tasks such as Visual
Question Answering (VQA), medical report generation task, which involve the
generation of longer sequences, are more prone to producing misleading or incorrect
information. This is due to the text-biased nature inherent in modern large
language models (MLLMs). An effective method for
estimating the uncertainty in the generated reports is to combine the model’s
capabilities in both captioning and VQA. As we have previously demonstrated the ability
to estimate uncertainty in VQA tasks,
we can apply the same approach to estimate the uncertainty of each sentence in the
generated medical report by transforming it into a VQA task.

Given a generated report $R = \{\vr_{1}, \vr_{2}, \ldots
, \vr_{n}\}$, where $\vr_{i}$ denotes the $i$-th generated sentence and $n$ is the total number of sentences, we estimate the uncertainty of
each sentence by generating $M$ VQA pairs for each sentence $\vr_{i}$.
These VQA pairs are related to the factual content of the sentence and are
represented as
$(\vx, \textbf{\text{q}}_{1}, \textbf{\text{a}}_{1}), (\vx
, \textbf{\text{q}}_{2}, \textbf{\text{a}}_{2}), \ldots, (\vx, \textbf{\text{q}}
_{M}, \textbf{\text{a}}_{M})$, where $\vx$ is the input image,
$\textbf{\text{q}}_{j}$ is the $j$-th question, and $\textbf{\text{a}}_{j}$ is the
corresponding answer. To simplify the process, we formulate the questions and answers
in a binary (yes/no) format. This binary structure enables more straightforward
comparison of generated answers with reference answers. For instance, for the
sentence "The patient has a mass in the right lung," the corresponding VQA question
might be ``Is there a mass in the right lung?” with the answer ``Yes.” For each sentence,
with $M$ VQA pairs, the model generates $M$ output sequences of tokens:
$\{\vs^{(1)}, \vs^{(2)}, \ldots, \vs^{(M)}
\}$, along with their associated token probabilities:
$\{P(\vs^{(1)}| \vx, \textbf{\text{q}}_{1}), P(\vs
^{(2)}| \vx, \textbf{\text{q}}_{2}), \ldots, P(\vs^{(M)}
| \vx, \textbf{\text{q}}_{M})\}$. Since the binary structure of the
questions and answers, the entailment between the generated answer and the
reference answer can be easily calculated by using rule-based methods and
clustering into two groups $C_{0}$ and $C_{1}$, where
$C_{0}$ denotes the group of non-semantically equivalent answers
and $C_{1}$ denotes the group of equivalent answers. The
semantic entropy $H$ is then calculated as:
\begin{equation}
    \centering
    H = P(C_{0}| \vx, \textbf{\text{q}})H_{C_0}+ P(
    C_{1}| \vx, \textbf{\text{q}})H_{C_1},
\end{equation}
where $H_{c_0}$ and $H_{c_1}$ are calculated by equation \Cref{eq:equation5}. The sentence-level
uncertainty estimation method is designed to provide a more accurate and robust
measure of uncertainty in the generated medical reports, and thus provide a more
verifiable and reliable tool for medical professionals to check and validate the
generated reports.

% attain $M$ visual-question-answering paris of $(\vx, \textbf{\text{q}
% }_{1}), (\vx, \textbf{\text{q} }_{2}), \ldots, (\vx,
% \textbf{\text{q}}_{M})$, where $\textbf{\text{x} }$ is the input image and
% $\textbf{\text{q}}_{j}$ is the $j$-th question related to the sentence
% $\textbf{\text{r}}_{i}$.

\subsection*{Formulation for medical reasoning}
\begin{figure}[t]
    \centering
    \includegraphics[width=1.0\linewidth]{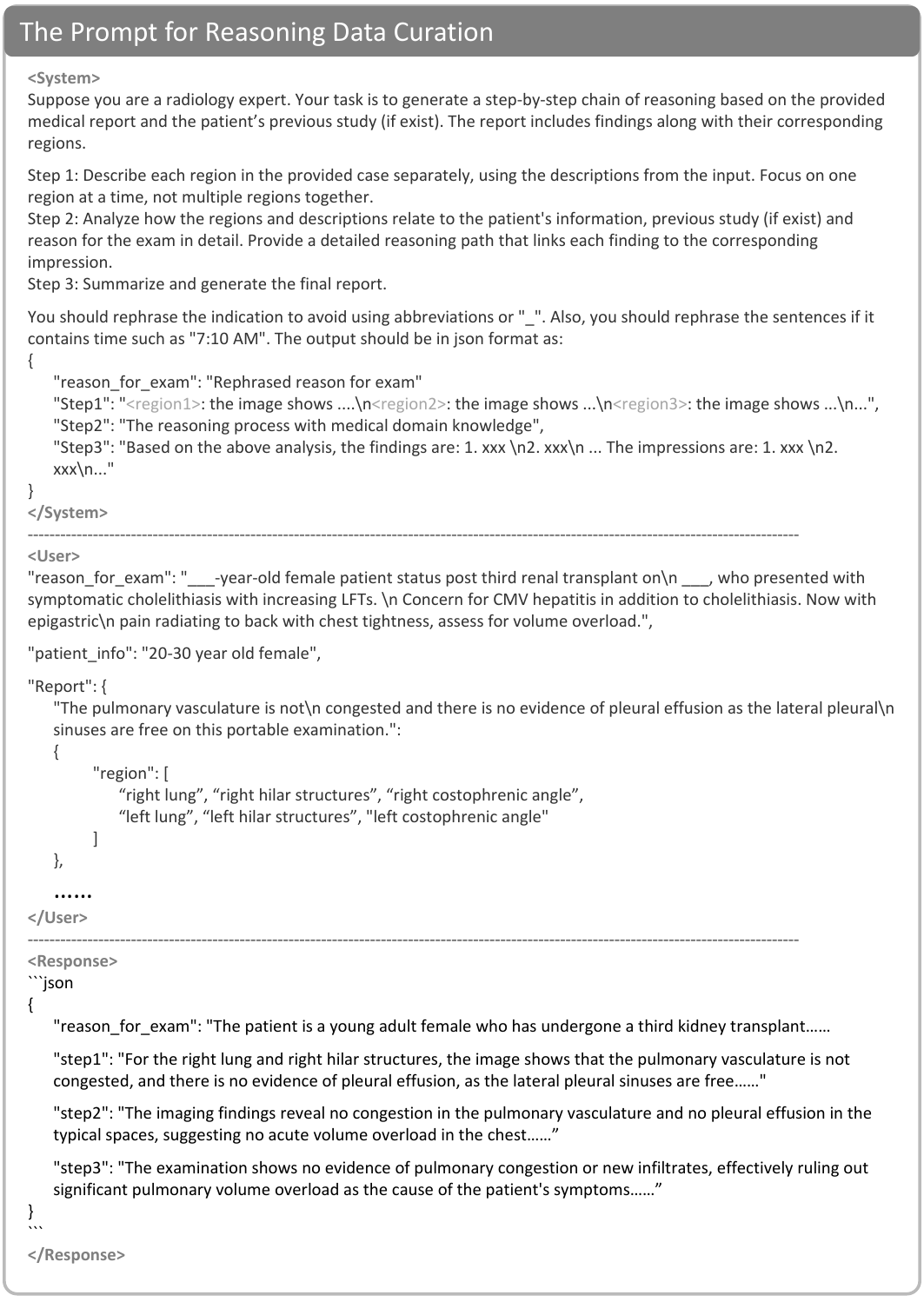}
    \caption{Illustration of the prompt design used for reasoning data curation.}
    \label{fig:Prompt_reasoning}
\end{figure}

\noindent
\textbf{Reasoning dataset curation.} We developed a step-by-step process for medical
report generation, based on common clinical workflows: (1) Identify the anatomical
structure to be examined, (2) Detect the regions of the anatomical structures, (3)
Describe each detected region in detail, (4) Conduct a comprehensive analysis of
all information, and (5) Summarize and generate the final report. To create a
medical reasoning dataset, we used the Chest-ImageNome dataset, which includes
aligned sentences and region-of-interest (ROI) annotations. The first step in our
process was to divide the medical reports into individual sentences and extract the
corresponding ROI annotations. Next, we rephrased the patient's chief complaint and
prior report, replacing abbreviations with their full forms and removing
irrelevant information such as dates. For example, "Eval post-BAL" was replaced
with "Evaluate after post-bronchoalveolar lavage." Using the aligned sentences,
ROI annotations, patient chief complaint, prior report, and reference report, we
employed GPT-4o to generate the reasoning process, moving from the findings section
to the impression section. The detailed prompt is presented in \Cref{fig:Prompt_reasoning}. The reasoning
process integrates the patient's previous conditions, current findings, and professional
knowledge. To ensure the accuracy and reliability of the generated reasoning, a team
of experienced radiologists reviewed and validated the output. Finally, we curated
a dataset containing 10,562 samples for medical reasoning, each comprising
comprehensive patient information and a detailed professional reasoning process.

\noindent
\textbf{Training protocol for medical reasoning.}
% DPO equation.
We first fine-tuned \methodname{} on the medical reasoning dataset using
supervised fine-tuning for 1 epoch. This process fosters the model's capbility to
generate comprehensive and accurate medical reasoning. To further enhance the
reasoning ability of \methodname{}, we employed Direct Policy Optimization (DPO)
for reinforcement learning. 
Following the process of DPO, we first curated 8,201 preference pairs $(\vx, \vs_{w}, \vs_{l})$, where $\vx$ is
the inputs, $\vs_{w}$ is the generated reasoning output with higher
preference and $\vs_{l}$ is the generated reasoning output with lower
preference. Specifically, given the input $\vx$ including the patient's
chief complaint, prior report, and chest x-ray image, we sampled
$K$ output sequences of tokens, $\{\vs^{(1)}, \vs^{(2)}, \ldots, \vs^{(K)}\}$. For each output, we extracted
the generated report in ``\textbf{Step 5}'' to calculate the preference score. The
preference score is calculated as the CheXpert F1 socre between the generated report
and the reference report. The preference score is then used to determine the
higher preference output $\vs_{w}$ and the lower preference output
$\vs_{l}$. Using the preference data and based on the Bradeley-Terry
model, the original objective for prior Reinforcement Learning from Human
Feedback (RLHF)
\begin{equation}
    \centering
    \mathcal{L}_{\text{RLHF}}= -\mathbb{E}_{\vx \sim \mathcal{D}, \vs \sim \pi_{\theta}(\vs|\vx)}\left
    [ r_{\phi}(\vx, \vs) - \beta \mathbb{D}_{\text{KL}}\left[ \pi_{\theta}(\vs|\vx) \parallel
    \pi_{\text{ref}}(\vs|\vx) \right] \right]
\end{equation}
is modified to
\begin{equation}
    \centering
    \mathcal{L}_{\text{DPO}}(\pi_{\theta}; \pi_{\text{ref}}) = -\mathbb{E}_{(\vx,
    \vs_{w}, \vs_{l}) \sim \mathcal{D}}\left[ \log \sigma \left( \beta \log \frac{\pi_{\theta}(\vs_{w}|
    \vx)}{\pi_{\text{ref}}(\vs_{w}| \vx)}- \beta \log \frac{\pi_{\theta}(\vs_{l}| \vx)}{\pi_{\text{ref}}(\vs_{l}|
    \vx)}\right) \right]
\end{equation}
where $\mathcal{D}$ denotes the empirical distribution over the constructed preference pairs, $\sigma(\cdot)$ is the sigmoid function, $\pi_{\theta}(\vs|\vx)$ is the policy
of the model, $\pi_{\text{ref}}(\vs|\vx)$ is the reference policy, and $\beta$ is the
hyperparameter that control the deviation from the base reference policy $\pi_{\text{ref}}
(\vs|\vx)$.

\subsection*{Medical-Omni post-training}
We first curated the Medical-Omni dataset, which consists of 514,668 samples
sampled from a pre-trained dataset. The text queries were then transformed into audio
using the SpeechT5 model. To increase the diversity of the audio, we employed the
SpeechT5 HiFi-GAN Vocoder, a model designed for voice conversion, to modify the audio
into different voice styles. This variation in audio styles ensures the model’s
ability to generalize across various voice inputs, which is crucial for its performance
in real-world applications. The post-training process was carried out on the combined
Medical-Omni dataset and the pre-trained dataset with text queries. The mixture
of these datasets helps improve the model’s generalization across different
input modalities. The post-training was performed for one epoch. During this
training, we froze the weights of the vision encoder, the vision MLP, and the language
model, updating only the weights of the audio encoder and the audio MLP. For the
audio encoder, we utilized a pre-trained model from Intern-Omni, which leverages
OpenAI's open-source Whisper-large-v3 model. This model has been trained on a
vast amount of audio data and is known for its robust capabilities in speech
recognition and translation.

\subsection*{Implementation details for external validation}
% External Validation dataset的介绍
% 数据来源，数据量，每个任务的数据情况
\noindent
\textbf{Dataset curation and evaluation tasks.} For a thorough external validation,
we collected CT scans and MRI scans that encompass a range of 11 different
diseases across three primary anatomical regions of the human body: HEAD and NECK
(including brain tumor, acoustic neuroma, and cerebral hemorrhage), CHEST (lung cancer,
pulmonary embolism), and ABDOMEN (bladder cancer, prostate cancer, colon cancer,
liver cancer, kidney cancer, pancreatic cancer). Each disease category includes
data from 20 patients, all sourced from Sun Yat-sen Memorial Hospital, Sun Yat-sen
University. The CT and MRI scans were meticulously annotated by experienced radiologists,
who provided detailed segmentation masks for each disease visible in the scans. For
the CT scans, we converted the DICOM images into PNG slices, adjusting the window
level and window width according to the radiologists' specifications. For the MRI
scans, we directly transformed the DICOM images into PNG slices without
additional adjustments. We then extracted the slices containing relevant disease
regions and corresponding bounding boxes for the purpose of external evaluation.
The final curated dataset consists of a total of 5,981
slices. The specific distribution of these slices for each disease is outlined in
\Cref{fig:intro}. This comprehensive dataset ensures a robust external validation of our
model across a wide range of diseases and anatomical regions.

The external validation process was carried out on four distinct tasks: (1) multiple-choice
question answering, (2) region recognition, (3) lesion detection, and (4) region-based
Visual Question Answering (VQA). For the region VQA task, we leveraged Llama-3.1-70B-Instruct~\cite{grattafiori2024llama3}
to extract sentences from medical reports that aligned with specific regions of
interest in the CT and MRI scans. These sentences were then used to construct VQA
pairs. In total, 2,760 region VQA test samples were curated for external validation.
For the multiple-choice question answering task, the model's performance was evaluated
by calculating the percentage of correct answers. The region recognition task
was assessed using the Macro-F1 score, which takes into account both precision and
recall to provide a balanced measure of the model's ability to correctly identify
regions of interest. The lesion detection task was evaluated using the
Intersection over Union (IoU) metric, which measures the overlap between the predicted
lesion regions and the ground truth annotations. For the region VQA task, the generated
answers were evaluated using GPT-4o. The evaluation involved scoring the model's
generated responses based on three key factors: (1) the accuracy of the answers,
(2) the consistency of the answers with the reference answers, and (3) the clinical
relevance of the answers, ensuring that the responses were both correct and
contextually meaningful in a clinical setting.

% Ablation studies, 分细粒度region的对比结果
% internal和external qualitative results
\noindent
\textbf{Implementation for Qualitative comparison of internal and external validation.} We
conducted a qualitative comparison of the internal and external validation
results to assess the model's generalization capabilities across different
datasets. The internal validation was performed on a held-out set
for region recognition and lesion detection. that contains 11 different diseases across four primary anatomical
regions: HEAD and NECK, CHEST, ABDOMEN, Pelvic. We compare the qualitative results
of region recognition and lesion detection tasks between the internal and
external validation datasets.

\section*{Code Availability}
The code used in this study will be publicly released after the manuscript is published. Implementation, including network design and training strategies, are given in the Methods section.

\section*{Data Availability}
This study incorporates diverse public available biomedical datasets. For detailed information about the data used in this project, please refer to the Appendix.

\clearpage
\newpage

\thispagestyle{empty}
\bibliographystyle{plainnat}
\bibliography{bib}

\clearpage
\newpage

\section*{Appendix}

\subsection*{Training Datasets}
\subsubsection*{Image Captioning}

\noindent\textbf{PMC-OA}~\cite{pmcoa} is a biomedical dataset with 1.6M image-caption pairs collected from PubMedCentral’s OpenAccess subset which covers diverse modalities or diseases. We leveraged the dataset for alignment training.

\noindent\textbf{QUILT}~\cite{quilt} is a large-scale vision-language dataset consisting of 802,144 image and text pairs, which was curated from video frames and corresponding subtitles on YouTube. We leveraged the dataset for alignment training.

\subsubsection*{Visual Question Answering}

\noindent\textbf{SLAKE}~\cite{liu2021slake} is a bilingual radiology VQA dataset consists of 642 radiology images and over 7000 diverse QA pairs annotated by experienced physicians. Following the official split, we used both English and Chinese versions for training, which contains 4919 and 4916 question-answer pairs repectively.

\noindent\textbf{VQA-RAD~}~\cite{vqarad} consists of 3.5K question-answering pairs on 314 radiology images, where clinicians asked naturally occurring questions about radiology images and provided reference answers. Following the official split, we use 3,064 question-answer pairs for training.

\noindent\textbf{PathVQA}~\cite{he2020pathvqa} consists of 32,799 open-ended questions from 4,998 pathology images where each question is manually checked to ensure correctness. Every image is paired with several questions related to multiple aspects such as shape, color and location. Following the official split, we use 19,755 question-answer pairs for training.

\noindent\textbf{PMC-VQA}~\cite{pmcvqa} consists of 1.6 million question-answer pairs, which is a large-scale medical visual question-answering dataset generated from PMC-OA. We combined two versions of PMC-VQA and use 329,551 question-answer pairs for training.

\noindent\textbf{MedPix} is collected from MedPix website\footnote{https://medpix.nlm.nih.gov}, which is a free open-access online database for medical usage. RadFM~\cite{wu2023radfm} separate the dataset into MPx-single and MPx-multi. We apply the MPx-single part and use 92,282 question-answer pairs for training.

\subsubsection*{Report Generation}

\noindent\textbf{MIMIC-CXR}~\cite{johnson2019mimiccxr} is a large-scale chest image-report dataset that consists of 371,920 chest X-rays associated with 227,943 reports from 65,079 patients. Following RadFM~\cite{wu2023radfm}, we use 354,569 cases for training. 

\noindent\textbf{IU-Xray}~\cite{demner2016iuxray} consists of 7,470 images and 3,955 reports collected from the Indiana Network. Following MedRegA~\cite{wang2024medrega}, we use 4,720 cases for training.

\subsubsection*{Medical Image Classification}

\noindent\textbf{VinDr-CXR}~\cite{nguyen2022vindrcxr} consists of 18,000 images that were manually annotated by a total of 17 experienced radiologists with 22 local labels of rectangles surrounding abnormalities and 6 global labels of suspected diseases. The training set contains 15,000 scans, and 3 radiologists independently label each image. Following the official split, we use 45,000 samples for training.

\noindent\textbf{VinDr-PCXR}~\cite{pham2022vindrpcxr} is a pediatric CXR dataset of 9,125 studies that were retrospectively collected from a major pediatric hospital in Vietnam between 2020-2021. Each scan was manually annotated by an experienced radiologist for the presence of 36 critical findings and 15 diseases. Following the official split, we use 4,585 samples for training.

\noindent\textbf{VinDr-SpineXR}~\cite{nguyen2021vindrspinexr} is a large-scale annotated medical image dataset for spinal lesion detection and classification from radiographs. The dataset contains 10,466 spine X-ray images from 5,000 studies, each of which is manually annotated with 13 types of abnormalities by an experienced radiologist with bounding boxes around abnormal findings. Following RadFM~\cite{wu2023radfm}, we use 8,389 samples for training.

\noindent\textbf{VinDr-Mammo}~\cite{nguyen2023vindrmammo} is a large-scale full-field digital mammography dataset of 5,000 four-view exams. Following the official split, we use 16,391 samples for training.

\noindent\textbf{CheXpert}~\cite{irvin2019chexpert} is a large public dataset for chest radiograph interpretation, which retrospectively collected the chest from Stanford Hospital, performed between October 2002 and July 2017. The dataset contains 224,316 chest radiographs of 65,240 patients. Following the official split, we use 223,414 samples for training.

\noindent\textbf{MURA}~\cite{rajpurkar2017mura} is a large-scale dataset of musculoskeletal radiographs containing 40,561 images from 14,863 studies, where each study is manually labeled by radiologists as either normal or abnormal. Following the official split, we use 36,808 samples for training.

\noindent\textbf{ISIC2018}~\cite{ISIC2018} is a skin lesion dataset acquired with 7 dermatoscope types. Following the official split, we use 10,015 samples for training.

\noindent\textbf{ISIC2019}~\cite{ISIC2019} is a skin lesion dataset labeled with 8 different categories. Following the official split, we use 25,331 samples for trianing.

\noindent\textbf{PAD-UFES}~\cite{pacheco2020padufes20} is a skin lesion dataset composed of clinical images collected from smartphone devices and a set of patient clinical data containing up to 22 features. The dataset consists of 1,373 patients, 1,641 skin lesions, and 2,298 images for six different diagnostics. We randomly sample 80\% of the dataset for training, which includes 1,838 samples.

\noindent\textbf{Kather colon dataset}~\cite{kather} is a dataset of 100,000 non-overlapping image patches from hematoxylin \& eosin (H\&E) stained histological images of human colorectal cancer (CRC) and normal tissue, covering 9 tissue classes in total.

\noindent\textbf{BRSET}~\cite{nakayama2024brset} is a multi-labeled ophthalmological dataset onsisting of 16,266 images from 8,524 Brazilian patients. Multi-labels are included alongside color fundus retinal photos. We randomly sample 80\% of the dataset for training, containing 13,012 samples.

\noindent\textbf{ODIR-5K} \footnote{https://odir2019.grand-challenge.org/} is a structured ophthalmic database of 5,000 patients with age, color fundus photographs from left and right eyes and doctors' diagnostic keywords from doctors. Following the official split, we use 6,392 samples for training.

\noindent\textbf{OCT2017} \footnote{https://www.kaggle.com/paultimothymooneyIkermany2018.} includes 83,484 OCT images of 4,686 patients, consisting of 4 categories, normal, drusen, choroidal neoVascularisation (CNV), and Diabetic Macular Edema (DME). Following the official split, we use 82,484 samples for training.

\noindent\textbf{Butterfly Network ultrasound dataset} \footnote{https://github.com/ButterflyNetwork/MITGrandHack2018} is a large dataset containing 9 different classes of ultrasound images acquired with the Butterfly IQ on 31 individuals. Following MedRegA~\cite{wang2024medrega}, 34,325 images are applied for training.

\noindent\textbf{BUSI}~\cite{busi} includes breast ultrasound images among women between 25 and 75 years old. The number of patients is 600 females, patients. The dataset consists of 780 images that are categorized into three classes, namely, standard, benign, and malignant. We randomly sample 80\% of the dataset for training, which includes 630 images.

\subsubsection*{Region-centric Dataset}
\noindent\textbf{SA-Med2D-20M}~\cite{SA-Med2D-20M} is a large-scale segmentation dataset of 2D medical images built upon numerous public and private datasets. The dataset consists of 4.6 million 2D medical images and 19.7 million corresponding masks, covering almost the whole body and showing significant diversity. We filter approximately 285K images from the original dataset, and construct 242,268 and 229,340 training samples for Region-to-Text Identification and Text-to-Region Detection respectively.

\noindent\textbf{VinDr Series Dataset} is a large-scale classification composed of VinDr-CXR, VinDr-PCXR, VinDr-SpineXR, VinDr-Mammo. The datasets provide radiologist’s bounding-box annotation for abnormal areas. We follow the official split and apply the samples with bounding boxes.

\noindent\textbf{ISIC Challenge Dataset} contains lesion segmentation data where the original image is paired with manually annotated lesion boundaries. We follow the official split and convert the segmentation map into bounding boxes.

\noindent\textbf{PanNuke}~\cite{pannuke} is a semi-automatically generated nuclei instance segmentation dataset. We follow the official split and convert the segmentation map into bounding boxes to formulate region-text pairs.

\noindent\textbf{Chest-ImaGenome}~\cite{Chest-ImaGenome} applied a CXR bounding box detection pipeline to automatically label frontal chest x-ray images from MIMIC-CXR dataset with 29 annotations, from which we selected 12 standardized structures in the chest. Following the split of MIMIC-CXR in RadFM~\cite{wu2023radfm}, we filtered the chext x-ray scans paired with annotation boxes, and obtained 222,588 samples for training.

\noindent\textbf{MM-WHS}~\cite{mmwhs} 
 (Multi-Modality Whole Heart Segmentation) is a challenge of MICCAI 2017 with a dataset of 120 multimodal cardiac images, including 60 cardiac CT/CTA and 60 cardiac MRI images.

\noindent\textbf{MSD Pancreas}~\cite{msdcolon_pancreas} aims to segment the pancreas and tumours from CT images. MSD chose this dataset due to label imbalance, which includes large (background), medium (pancreas), and small (tumour) structures. The dataset comprises 420 cases of 3D CT data, officially split into 281 training cases and 139 test cases. This dataset includes three types of pancreatic tumours: intraductal papillary mucinous neoplasm, pancreatic neuroendocrine tumour, and pancreatic ductal adenocarcinoma.

\noindent\textbf{MSD Colon Cancer }~\cite{msdcolon_pancreas} aims to segment colon tumours from CT images. The reason MSD chose this dataset is due to the challenges posed by heterogeneous appearances and the difficulties in annotation. The dataset includes portal vein CT scans of 190 patients who underwent resection surgery for primary colon cancer, and is officially divided into 126 training 
cases and 64 testing cases.

\noindent\textbf{GAMMA}~\cite{gamma} consists of 2D fundus images and 3D optical coherence tomography (OCT) images of 300 patients. The dataset was annotated with glaucoma grade in every sample, and macular fovea coordinates as well as optic disc/cup segmentation mask in the fundus image.

\noindent\textbf{PathMNIST}~\cite{yang2023medmnist} is for predicting survival from colorectal cancer histology slides, providing a dataset
 (NCT-CRC-HE-100K) of 100000 non-overlapping image patches from hematoxylin \& eosin stained histological images, and
 a test dataset (CRC-VAL-HE-7K) of 7180 image patches from a different clinical center.

\noindent\textbf{ChestMNIST}~\cite{yang2023medmnist} is a dataset comprising 112120 frontal-view X-Ray images
 of 30805 unique patients with the text-mined 14 disease labels, which could be formulized as a multi-label binary-class
 classification task. We use the official data split, and resize the source images of 1$\times$1024$\times$1024 into 1$\times$28 $\times$28.
 
\noindent\textbf{DermaMNIST}~\cite{yang2023medmnist} consists of 10015 dermatoscopic images categorized as 7 different diseases, formulized as
 a multi-class classification task. We split the images into training, validation and test set with a ratio of 7 : 1 : 2. The source
 images of 3$\times$600$\times$450are resized into 3$\times$28$\times$28.
 
\noindent\textbf{OCTMNIST}~\cite{yang2023medmnist}  is comprised of 4 diagnosis categories, leading to a multi-class classification task. We split the source
 training set with a ratio of 9 : 1 into training and validation set, and use its source validation set as the test set. The source
 images are gray-scale, and their sizes are (384-1536) (277-512). We center-crop the images with a window size of
 length of the short edge and resize them into 1$\times$28$\times$28.
 
\noindent\textbf{Organ{A,C,S}MNIST}~\cite{yang2023medmnist} is based on 3D computed tomography (CT) images from Liver Tumor Segmentation Benchmark. They are renamed from OrganMNIST(in MedMNIST v19) for simplicity. We use bounding
box annotations of 11 body organs from another study30 to obtain the organ labels. Hounsfield-Unit (HU) of the 3D images are
 transformed into gray-scale with an abdominal window. We crop 2D images from the center slices of the 3D bounding boxes in
 axial / coronal / sagittal views (planes). The only differences of Organ{A,C,S}MNIST are the views. The images are resized
 into 1 28 28toperform multi-class classification of 11 body organs. 115 and 16 CT scans from the source training set are
 used as training and validation set, respectively. The 70 CT scans from the source test set are treated as the test set.
 
\noindent\textbf{PneumoniaMNIST}~\cite{yang2023medmnist} consists 5856 pediatric chest X-Ray images. The task is binary-class
 classification of pneumonia against normal. We split the source training set with a ratio of 9 : 1 into training and validation set,
 and use its source validation set as the test set. The source images are gray-scale, and their sizes are (384-2916) (127-2713). We center-crop the images with a window size of length of the short edge and resize them into 1$\times$28$\times$28.
 
\noindent\textbf{RetinaMNIST}~\cite{yang2023medmnist} provides a dataset of 1600 retina fundus images. The task
 is ordinal regression for 5-level grading of diabetic retinopathy severity. We split the source training set with a ratio of 9 : 1
 into training and validation set, and use the source validation set as the test set. The source images of 3 1736 1824 are
 center-cropped with a window size of length of the short edge and resized into 3$\times$28$\times$28.
 
\noindent\textbf{BreastMNIST}~\cite{yang2023medmnist}  is categorized into 3 classes: normal, benign, and
 malignant. As we use low-resolution images, we simplify the task into binary classification by combining normal and benign as
 positive and classifying them against malignant as negative. We split the source dataset with a ratio of 7 : 1 : 2 into training,
 validation and test set. The source images of 1 500 500 are resized into 1$\times$28$\times$28
 
\noindent\textbf{BloodMNIST}~\cite{yang2023medmnist} is a dataset of individual normal cells, captured from individuals without infection, hematologic
 or oncologic disease and free of any pharmacologic treatment at the moment of blood collection. It contains a total of 17,092
 images and is organized into 8 classes. We split the source dataset with a ratio of 7 : 1 : 2 into training, validation and test
 set. The source images with resolution 3$\times$360$\times$360 pixels are center-cropped into 3 200 200, and then resized into
 3$\times$28$\times$28.
 
\noindent\textbf{TissueMNIST}~\cite{yang2023medmnist}  contains 236386 human
 kidney cortex cells, segmented from 3 reference tissue specimens and organized into 8 categories. We split the source dataset
 with a ratio of 7 : 1 : 2 into training, validation and test set. Each gray-scale image is 32$\times$32$\times$7 pixels, where 7 denotes 7
 slices. We obtain 2D maximum projections by taking the maximum pixel value along the axial-axis of each pixel, and resize
 them into 28$\times$28 gray-scale images.
 
\noindent\textbf{BTCV}\footnote{https://www.synapse.org/Synapse:syn3193805/wiki/89480} A total of 50 abdominal CT scans were included, which were derived from patients with metastatic liver cancer or postoperative abdominal wall hernias. Each scan of the dataset was performed during the portal vein contrast phase with different volume and field of view parameters. The in-plot resolution in the dataset varies from 0.54 x 0.54 mm² to 0.98 x 0.98 mm², and the slice thickness ranges from 2.5 mm to 5.0 mm.

\noindent\textbf{AutoPET}~\cite{AutoPET} is a large-scale PET/CT image dataset focused on whole-body tumour segmentation. The training set consisted of 1014 paired PET-CT images of 900 patients, and the test set consisted of 200 images. Each case data consists of a 3D whole-body FDG-PET image, a 3D whole-body CT image, and a manually annotated 3D tumour mask.

\noindent\textbf{Knee Osteoarthritis}~\cite{KneeOsteoarthritis} contains knee X-ray data for knee testing and Kellgren–Lawrence grading. A total of 9786 images of the knee are classified into 5 severity levels according to the Kellgren–Lawrence (KL) grading system: 0 (healthy), 1 (suspicious), 2 (mild), 3 (moderate), and 4 (severe). All images have a resolution of 224 × 224 pixels. About 40\% of the dataset images fall into the health category, while the proportion of suspicious images is about 18\%, the proportion of minimal images is 26\%, the proportion of medium images is 13\%, and the proportion of severe images is just over 3\%.

\noindent\textbf{CHAOS}~\cite{chaos} is one of the classic benchmarks for abdominal medical image segmentation, and its biggest feature is that it provides pairs of multimodal CT and MR data and provides corresponding annotations. The CHAOS dataset was released in the ISBI 2019 Challenge, with a total of 40 paired CT and MR data, of which only 20 were annotated as training sets.

\noindent\textbf{MSD Lung Tumours}~\cite{MSDLungTumours} aims to segment lung tumours from CT images, and MSD chose this dataset because of "segmenting small targets in a large background". The dataset contains thin-slice CT scans of 96 patients with non-small cell lung cancer, which is officially divided into 64 training sets and 32 test sets.

\noindent\textbf{ChestX-ray14}  is a medical imaging dataset that contains 112,120 frontal view X-ray images from 30,805 patients, collected from 1992 to 2015. The dataset extracted labels for 14 common diseases from radiology reports using Natural Language Processing (NLP) techniques, including atelectasis, infiltration, pneumothorax, edema, emphysema, fibrosis, effusion, pneumonia, pleural thickening, cardiomegaly, nodules, masses, and hernia.

\noindent\textbf{AMOS}~\cite{amos} provides 500 CT and 100 MR scans from multi-centre, multi-vendor, multi-modal, multi-period, multi-disease patients, each with voxel-level annotations for 15 abdominal organs.Compared with the previous medical image analysis datasets, AMOS provides fine annotation of up to 15 abdominal organs while ensuring a large amount of data, which has high clinical value.

\noindent\textbf{EMIDEC}~\cite{emidec} is a dataset designed to assess the extent of myocardial infarction. The dataset collected delayed-enhanced magnetic resonance imaging (DE-MRI) images of multiple patients a few minutes after contrast injection, and manually annotated multiple myocardial infarction-related regions, including myocardial contours, infarct areas, and permanent microvascular occlusion areas (areas without reflux), to form a segmented dataset. The dataset contains data on 150 patients (all from different patients), of which 50 were normal MRI imaging after contrast media injection and 100 were myocardial infarction (showing areas of high enhancement on DE-MRI). The dataset contains 100 training sets and 50 test sets.

\noindent\textbf{Kvasir-SEG }~\cite{kvasir} is an endoscopic dataset for pixel-level segmentation of colon polyps, which includes 1000 images of gastrointestinal polyps and their corresponding segmentation masks, which are personally annotated and verified by senior gastroenterologists. The official data warehouse provides a breakdown of training and validation data at a ratio of 880:120.

\noindent\textbf{MSD Hepatic Vessel}~\cite{MSDLungTumours} aims to segment liver vessels and tumours from liver CT, and MSD chose this dataset because of "the nature of hepatic blood vessels that are adjacent, tubular and interconnected to heterogeneous tumours." The dataset contains 443 3D CT data, divided into 303 training sets and 140 test sets.

\noindent\textbf{WORD}~\cite{word} is a large-scale abdominal organ segmentation CT dataset. The dataset includes 150 CT scans that provide comprehensive coverage of the abdominal region and provide detailed annotations for 16 different abdominal organs. Officially, this data is divided into 100 for training, 20 for validation, and 30 for testing.

\noindent\textbf{KiTS23}~\cite{kits21} focuses on CT segmentation of the kidney and tumours and cysts on it. The dataset includes 599 cases, of which 489 were used for training and 110 for testing.

\noindent\textbf{OCTUD}~\cite{octid} contains over 500 high-resolution images categorized into different pathological conditions. The image classes include Normal (NO), Macular Hole (MH), Age-related Macular Degeneration (AMD), Central Serous Retinopathy (CSR), and Diabetic Retinopathy (DR).

\noindent\textbf{ICIAR}~\cite{iciar2018} contains a total of 400 microscope images distributed as follows: normal: 100, benign: 100, carcinoma in situ: 100, invasive carcinoma: 100 and performs two tasks: Automatically classify H\&E-stained breast histology microscopy images into four categories: normal, benign, carcinoma in situ, and invasive carcinoma and Perform pixel-level tagging on the entire slice image in the same four categories..

\noindent\textbf{VQA-Med}~\cite{ben2019medvqa}contains four main question categories: Modality, Plane, Organ System, and Abnormality. It contains a total of 4,200 radiology images and their corresponding 15,292 question-answer pairs, divided into a training set of 3,200 images and 12,792 question-answer pairs, a validation set of 500 images and 2,000 question-answer pairs, and a test set of 500 images and 500 questions.

\noindent\textbf{CrossMoDA}~\cite{crossmoda} is a large and multi-class benchmark for unsupervised cross-modality Domain Adaptation. The goal of the challenge is to segment two key brain structures involved in the follow-up and treatment planning of vestibular schwannoma (VS): the VS and the cochleas. Currently, the diagnosis and surveillance in patients with VS are commonly performed using contrast-enhanced T1 (ceT1) MR imaging.

\noindent\textbf{BRATS 2013}~\cite{brats2013} is a brain tumor segmentation dataset consists of synthetic and real images, where each of them is further divided into high-grade gliomas (HG) and low-grade gliomas (LG). There are 25 patients with both synthetic HG and LG images and 20 patients with real HG and 10 patients with real LG images.

\noindent\textbf{Atria 2018 }~\cite{zhang2023multi_Atria} is a dataset of 3D Gadolinium-Enhanced Magnetic Resonance Imaging, and contains 100 Data for Training and 54 Data for Testing.

\noindent\textbf{RFMiD }~\cite{panchal2023rfmid} is a multi-label classification dataset for fundus images. The dataset contains 3200 cases and provides annotations for 45 eye diseases.

\noindent\textbf{COSMOS} \footnote{https://vessel-wall-segmentation-2022.grand-challenge.org/} is a dataset designed to segment the blood vessel wall from the influence of 3D-VISTA and accurately diagnose atherosclerotic lesions. The dataset provides 50 cases for model training and validation, and 25 cases for testing.

\noindent\textbf{CT-ORG}~\cite{CTORG} is a CT multi-organ segmentation dataset that officially provides annotations of 6 organs: lungs, bones, liver, kidneys, bladder, and brain, including 140 CT scans, of which 131 were directly derived from LiTS and 9 were derived from CT data from PET-CT at Harvard Medical School. Officially, the dataset is divided into 119 training sets and 21 test sets.

\noindent\textbf{PROMISE12}~\cite{PROMISE12} is a classic dataset in the field of medical image segmentation, and as part of the MICCAI 2012 challenge, 50 prostate MRI images and their corresponding segmentation annotations were provided.

\noindent\textbf{TotalSegmentator}~\cite{totalsegmentator} is currently the largest publicly available dataset in the field of 3D medical image segmentation, including 1204 CT images covering 104 anatomical structures throughout the body. Of these, 1082 were used for training, 57 for validation, and 65 for the test set.

\noindent\textbf{BHX}~\cite{bhx} is a public available dataset with bounding box annotations for 5 types of acute hemorrhage as an extension of the qure.ai CQ500 dataset, containing up to 39,668 bounding boxes in 23,409 images annotated for hemorrhage, out of a total of ~170k images from qure.ai CQ500 dataset.

\noindent\textbf{MSD Prostate}~\cite{MSDLungTumours} aims to segment two regions of the prostate, the central gland and the peripheral band, from multimodal MR (T2, ADC) images. The dataset includes T2 and ADC MR images of 48 patients, which are officially divided into 32 training sets and 16 test sets

\noindent\textbf{The Bone Marrow Cytomorphology}~\cite{TheBoneMarrowCytomorphology} is a dataset focused on bone marrow cell morphology, contains more than 170,000 de-identified, expert-annotated images of cells from bone marrow smears from 945 patients using May-Grünwald-Giemsa/Pappenheim staining

\noindent\textbf{PALM19}~\cite{PALM} is focusing on the research and development of advanced algorithms for the diagnosis of pathological myopia (PM), as well as the accurate segmentation of lesion areas in the fundus photos of PM patients, a total of 1,200 images with detailed annotations were provided, which were divided into 400 training images, 400 validation images and 400 test images to support the algorithm development and testing of the participants.

\noindent\textbf{OmniMedVQA}~\cite{hu2024omnimedvqa} is collected from 73 different medical datasets, including 12 different modalities and covering more than 20 distinct anatomical regions. Importantly, all images in this benchmark are sourced from authentic medical scenarios, ensuring alignment with the requirements of the medical field and suitability for evaluating LVLMs. 

\noindent\textbf{ISLES22 }~\cite{isles} is designed to automatically segment acute to subacute ischemic stroke lesions by multimodal MR images, including FLAIR, DWI, and ADC. The dataset brings together 400 multicenter, multi-device MRI cases that exhibit a high degree of variability in the size, number, and location of stroke lesions. The dataset is divided into a training set of 250 cases and a test set of 150 cases.

\noindent\textbf{FUMPE}~\cite{FUMPE}is a dataset for segmentation of pulmonary embolism (PE) in computed tomography (CTA) images, containing images of 35 different patients.

\noindent\textbf MSD Cardiac~\cite{MSDLungTumours} aims to segment the left atrium from unimodal MR images, and the dataset includes MR images of 30 patients, which are officially divided into 20 training sets and 10 test sets,

\noindent\textbf{Med-GRIT-270k}~\cite{huang2024bird} is a large-scale biomedical image-mask pairs are transformed into multi-modal conversations by leveraging chatGPT [19] in a novel process. It is the first dataset in biomedicine to integrate referring, grounding, and conversations.

\noindent\textbf{PitVis }~\cite{das2024pitvis} is a dataset of endoscopic pituitary surgery videos in PitVis, and provide live annotations of steps and instrument grounds, including 120024 images, and randomly divide the training set and validation set at a scale of 8:2

\noindent\textbf{RP3D}~\cite{wu2023radfm} has four subsets of Caption/VQA/Rationale/Modality, which share the same data source from Radiopaedia, a medical imaging teaching website that contains 3D radiography images uploaded by doctors and descriptions.

\noindent\textbf{BTCV Cervix}~\cite{BTCV-Cervix}  is a CT segmentation dataset for cervical cancer patients that provides radiation therapy planning, annotating 4 organs of the bladder, uterus, rectum and small intestine, containing 30 annotated training data and 20 unannotated test data.

\noindent\textbf{CMRxMotion}~\cite{CMRxMotion} is designed to explore cardiac MRI analysis in extreme conditions of respiratory motion. It consists of two tasks: 1) image quality evaluation and 2) segmentation. The challenge consists of 160 training sets, 40 validation sets, and 160 test sets.

\noindent\textbf{SIIM-FISABIO-RSNA COVID-19}~\cite{siim} is a dataset focused on the use of advanced machine learning techniques to accurately identify and locate COVID-19 in chest radiographs. The dataset includes 6,334 high-quality chest X-rays in DICOM format.

\subsection*{Test Datatsets}
\subsubsection*{Visual Question Answering}
\noindent\textbf{SLAKE}. Following the official split, we use 1,061 and 1,033 quesion-answer pairs for test on the English and Chinese version, respectively.

\noindent\textbf{VQA-RAD}. Following the official split, we use 451 quesion-answer pairs for evaluation.

\noindent\textbf{PathVQA}. Following the official split, we use 6,761 quesion-answer pairs for evaluation.

% \noindent\textbf{PMC-VQA}. Following the official split, the two versions of PMC-VQA test set contain 50,000 and 33,430 quesion-answer pairs, respectively. We test the model on both versions and report the averaged result.

\noindent\textbf{OmnimedVQA}. We used the open access part of the OmniMedVQA dataset, which includes 42
traditional medical imaging datasets, all formatted as VQA.

\noindent\textbf{GMAI-MMbench}. The most comprehensive general medical AI benchmark with well-categorized data structure and multi-perceptual granularity to date. It contains 21,281 test data constructed from 284 datasets across 38 medical image modalities, 18 clinical-related tasks, 18 departments, and 4 perceptual granularities in a Visual Question Answering (VQA) format.

\subsubsection*{Report Generation}
\noindent\textbf{MIMIC-CXR}. Following RadFM~\cite{wu2023radfm}, we use 4,710 cases for test. 

\noindent\textbf{IU-Xray}. Following MedRegA~\cite{wang2024medrega}, we use 1,180 cases for test.

\subsubsection*{Medical Image Classification}
\noindent\textbf{MURA}. Following the official split, we use 3,193 X-ray images for test.

\noindent\textbf{PAD-UFES-20}. Since no official split is provided, we randomly split 20\% of the data for evaluation, which contains 1,044 samples.

\noindent\textbf{BRSET}. Since no official split is provided, we randomly split 20\% of the data for test, which includes 3,254 fundus images.

\noindent\textbf{RFMiD 1.0}. Following the official split, we use 640 fundus images for evaluation.

\noindent\textbf{PneumoniaMNIST}. Following the official split, we use 624 X-ray samples for test.

\noindent\textbf{OrganCMNIST}. Following the official split, we use 8,216 CT samples for test.

\noindent\textbf{OrganAMNIST}. Following the official split, we use 17,778 CT samples for test.

\noindent\textbf{OrganSMNIST}. Following the official split, we use 8,827 CT samples for test.

\noindent\textbf{OCTMNIST}. Following the official split, we use 1,000 OCT scans for test.

\noindent\textbf{BreastMNIST}. Following the official split, we use 156 ultrasound scans for test.

\noindent\textbf{ChestMNIST}. Following the official split, we use 22,433 chest x-ray images for test.

\noindent\textbf{PathMNIST}. Following the official split, we use 7,179 pathology samples for test.

\noindent\textbf{VinDr-CXR}. Following the official split, we use 3,000 chest x-ray images for evaluation.

\noindent\textbf{VinDr-PCXR}. Following the official split, we use 1,387 chest x-ray images for evaluation.

\noindent\textbf{VinDr-SpineXR}. Following the official split, we use 2,077 spine x-ray images for evaluation.

\noindent\textbf{VinDr-Mammo}. Following the official split, we use 4,000 mammography images for evaluation.

\subsubsection*{Lesion Detection and Region Recognition}
"We evaluate our model using 1,272 test samples for region recognition and 1,000 test samples for lesion detection, each spanning 141 regions. These samples are selected from the held-out test set of the SA-Med-2D dataset. Additionally, we assess our model on the MedGRIT benchmarks, adhering to the conventions and metrics established in BiRD~\cite{huang2024bird}. MedGRIT consists of four tasks: (i) Region-in and Text-out tasks, which include Referring Object Classification (ROC) with 5,293 samples and Referring Captioning (RC) with 5,265 samples; (ii) Text-in and Region-out tasks, which involve Visual Grounding (VG) with 5,276 samples; and (iii) Text-in and Text-out tasks, represented by Medical Image Analysis (MIA) with 5,293 samples.

\subsubsection*{External Validation}
We collected CT and MRI scans covering 11 distinct diseases across three major anatomical regions: the HEAD and NECK (including brain tumors, acoustic neuromas, and cerebral hemorrhage), CHEST (lung cancer and pulmonary embolism), and ABDOMEN (bladder, prostate, colon, liver, kidney, and pancreatic cancers). Each disease category comprises data from 20 patients, all obtained from Sun Yat-sen Memorial Hospital, Sun Yat-sen University. In total, the dataset includes 5,981 slices for external validation.

\subsection*{Prognostic Prediction}

\noindent\textbf{Lung Tumor Progression-Free Survival Prediction}. The model was trained on a multicenter dataset comprising patients from two hospitals: ZE (n = 238) and NF (n = 85). For external validation, it was evaluated on independent datasets from two additional hospitals: JX (n = 58) and HN (n = 49). During training, the data were split into an 8:2 ratio for training and internal evaluation, respectively.

\noindent\textbf{NSCLC Overall Survival Prediction}. The model was trained on the Lung1 dataset (n = 420) and externally validated on the RADIO cohort (n = 133). The training data were divided into an 8:2 ratio for training and internal evaluation.

\noindent\textbf{Glioma Tumor Overall Survival Prediction}. The model was trained on the UPENN-GBM cohort (n = 585) and externally validated on the BraTS dataset (n = 163). The training data were split into an 8:2 ratio for training and internal evaluation.

\clearpage
\newpage

\begin{figure}[t]
    \centering
    \includegraphics[width=1.0\linewidth]{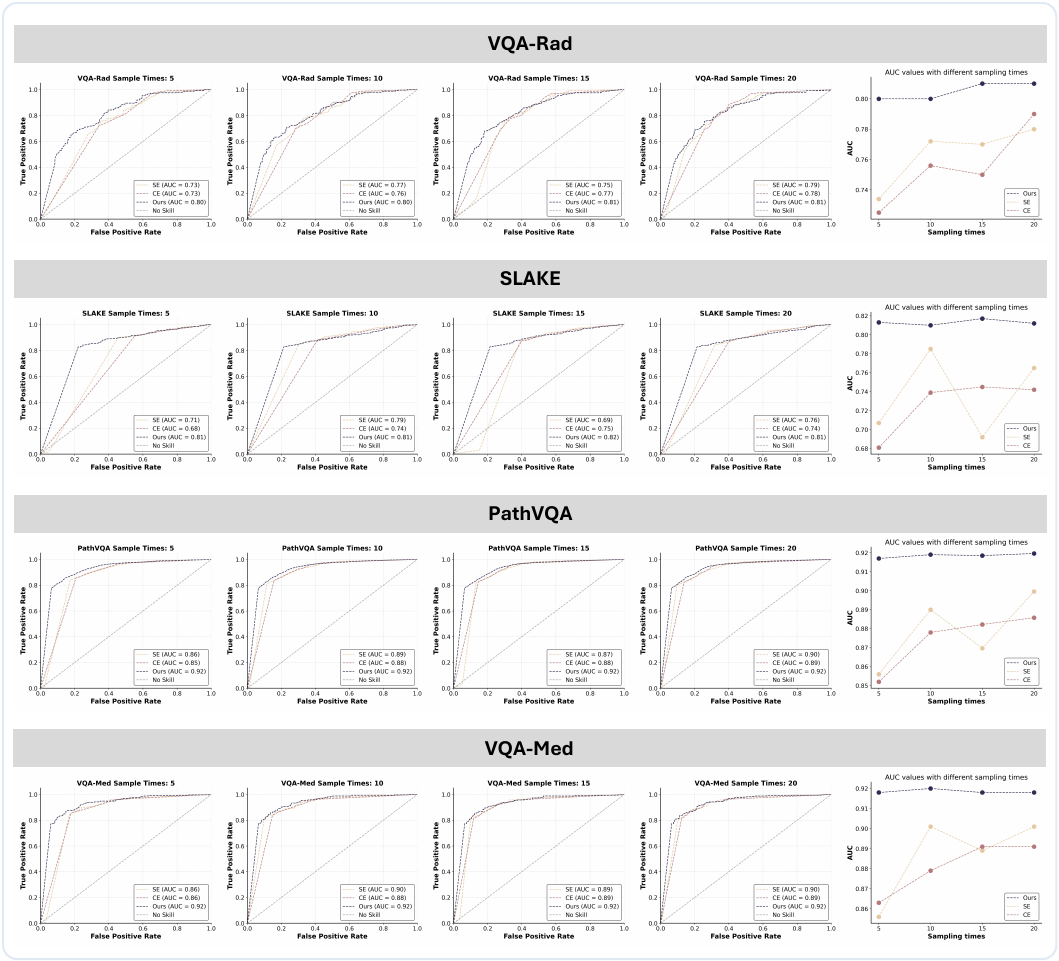}
    \caption{Ablation results on four VQA datasets demonstrating the impact of sampling frequency on uncertainty estimation. Our method consistently outperforms conventional semantic entropy, particularly when the number of sampling iterations is fewer than 10, and achieves higher AUC scores as sampling increases.}
    \label{fig:Uncertainty_ablation}
\end{figure}

\begin{figure}[t]
    \centering
    \includegraphics[width=1.0\linewidth]{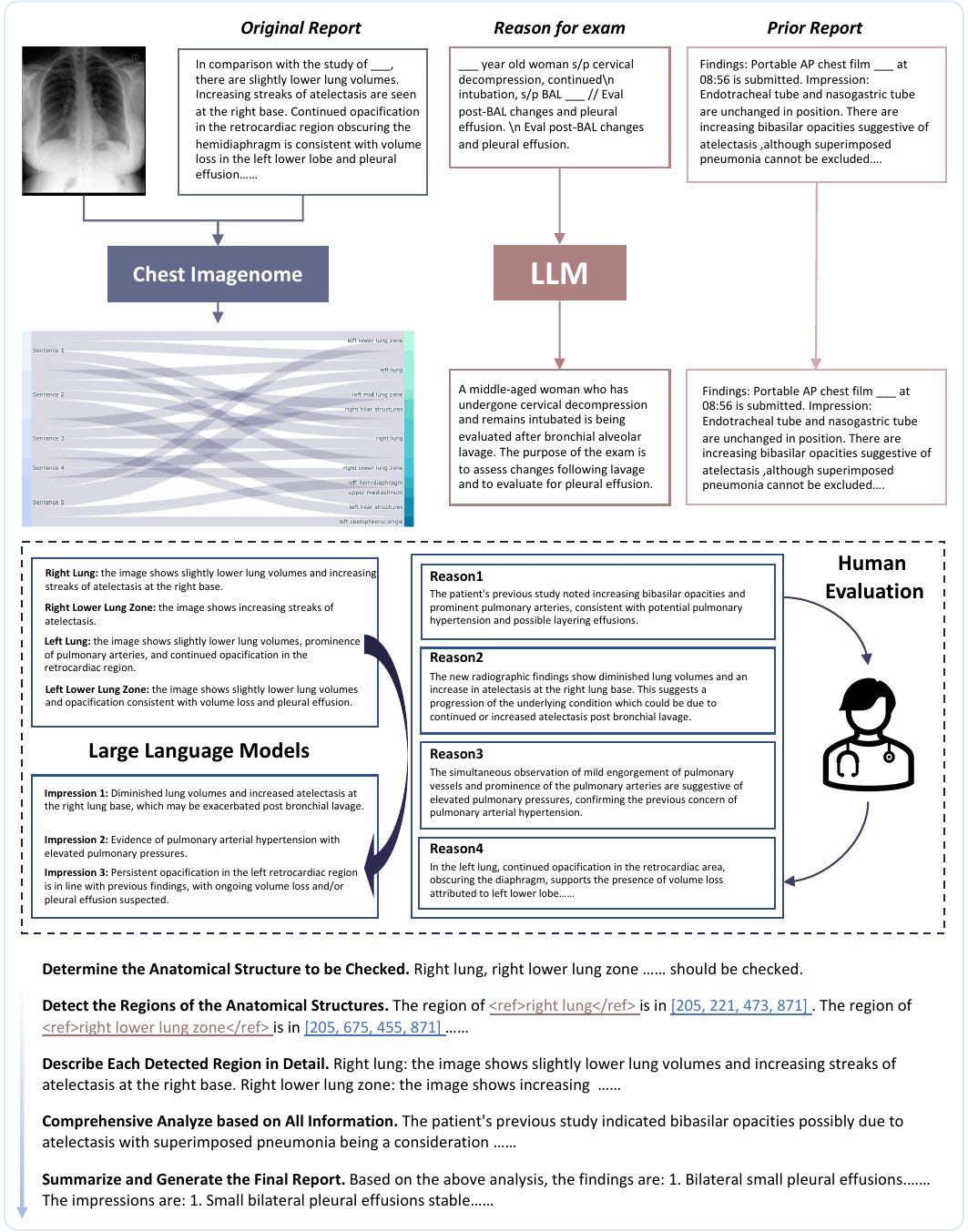}
    \caption{Schematic overview of the data curation pipeline for reasoning tasks.}
    \label{fig:Reasoning_data}
\end{figure}

\begin{figure}[t]
    \centering
    \includegraphics[width=1.0\linewidth]{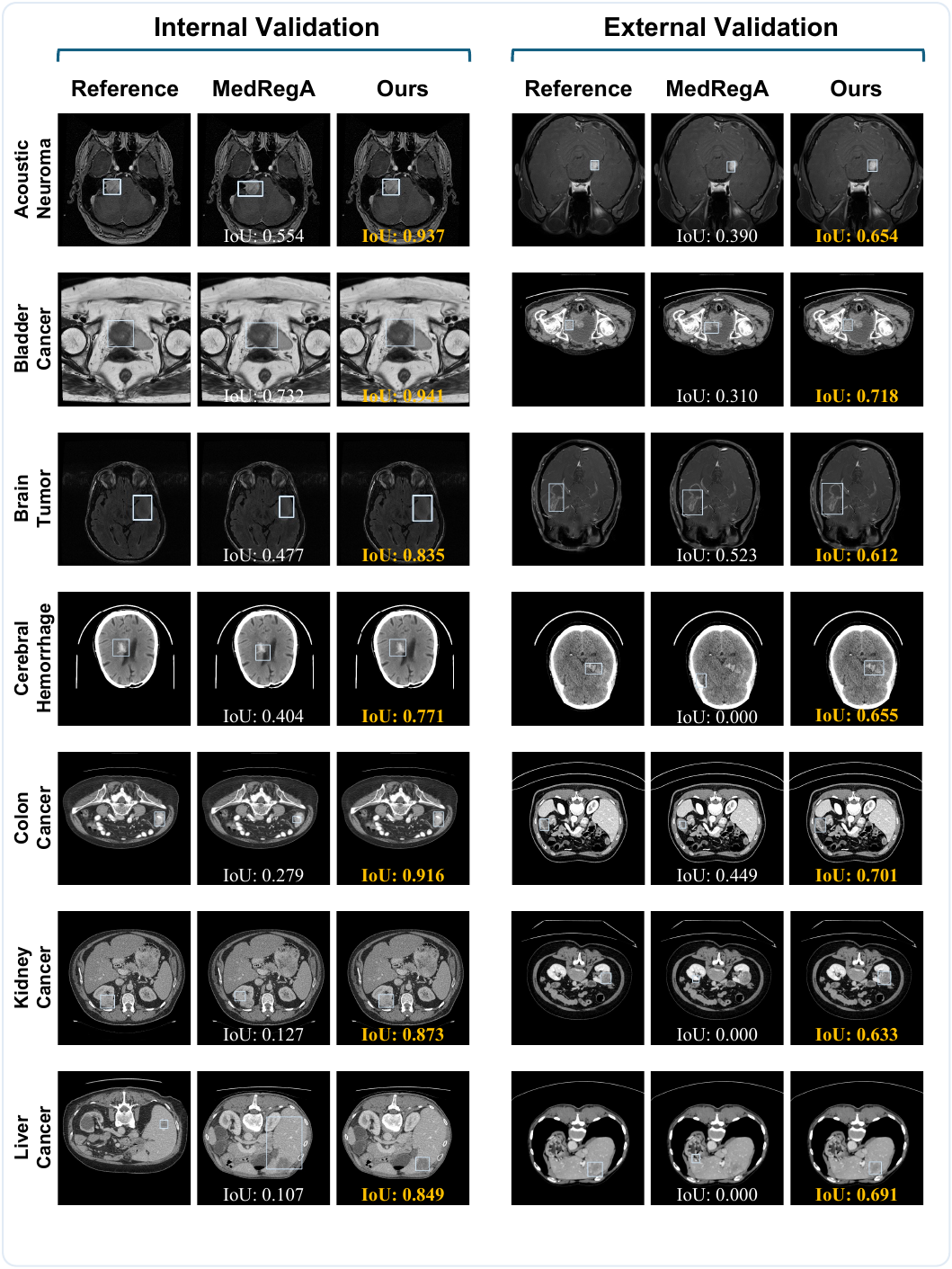}
    \caption{Qualitative comparison between XmedGPT and a state-of-the-art Generalist Medical AI (GMAI) model on lesion detection tasks.}
    \label{fig:Det_results1}
\end{figure}

\begin{figure}[t]
    \centering
    \includegraphics[width=1.0\linewidth]{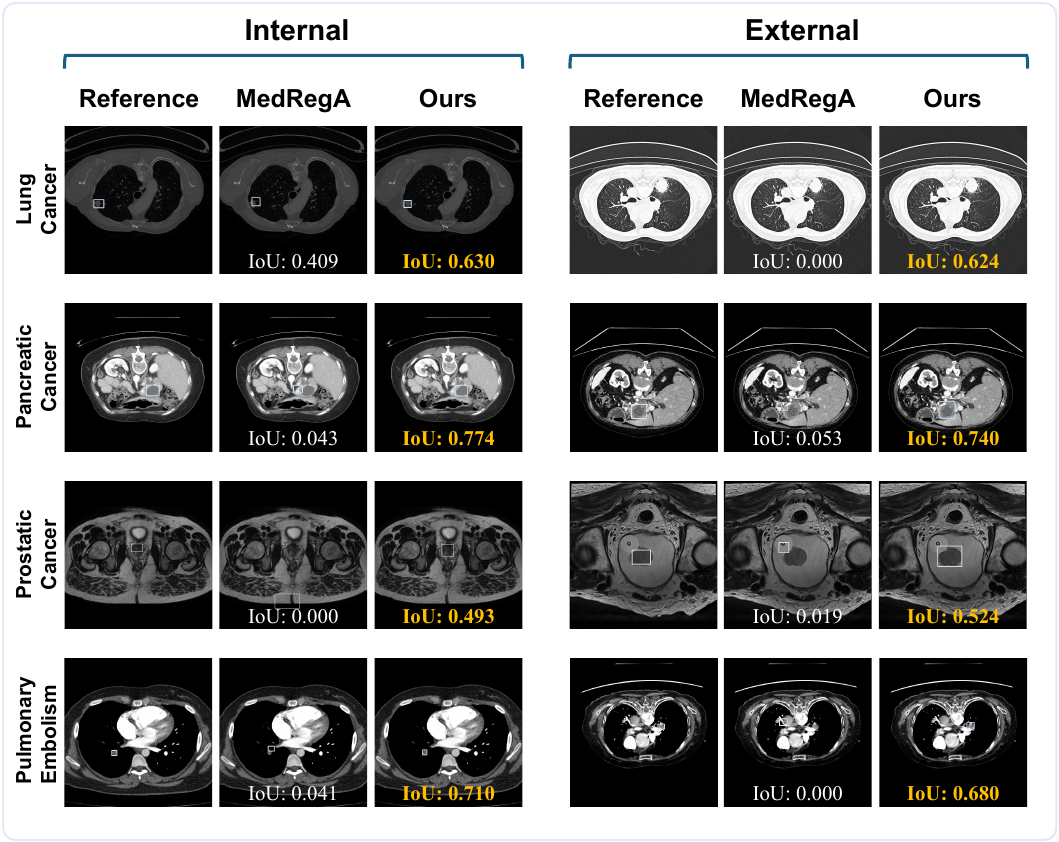}
    \caption{Qualitative comparison between XmedGPT and a state-of-the-art Generalist Medical AI (GMAI) model on lesion detection tasks.}
    \label{fig:Det_results2}
\end{figure}

\begin{figure}[t]
    \centering
    \includegraphics[width=1.0\linewidth]{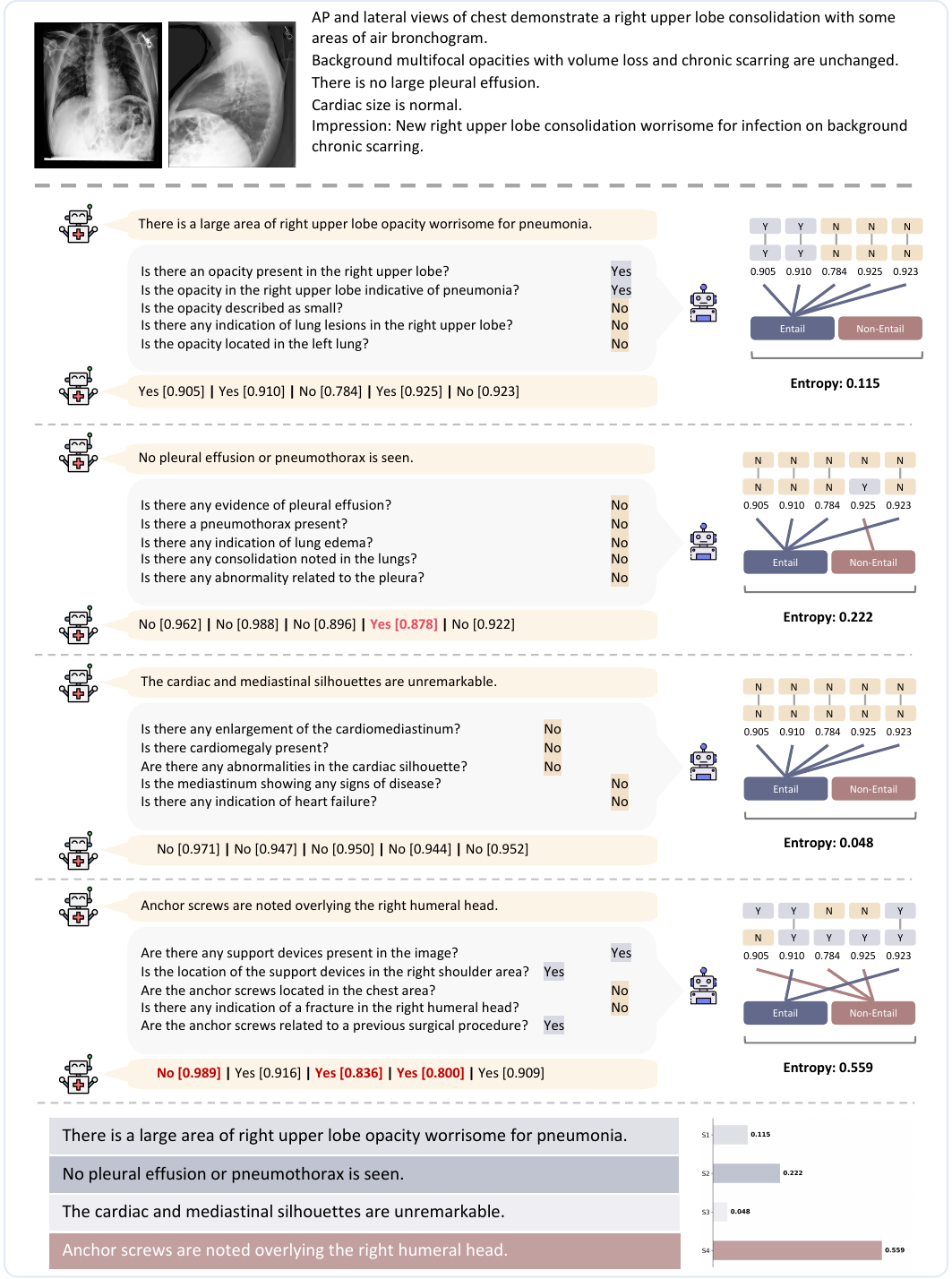}
    \caption{Representative example illustrating the uncertainty estimation pipeline and its corresponding results.}
    \label{fig:Uncertainty_result}
\end{figure}

\begin{figure}[t]
    \centering
    \includegraphics[width=1.0\linewidth]{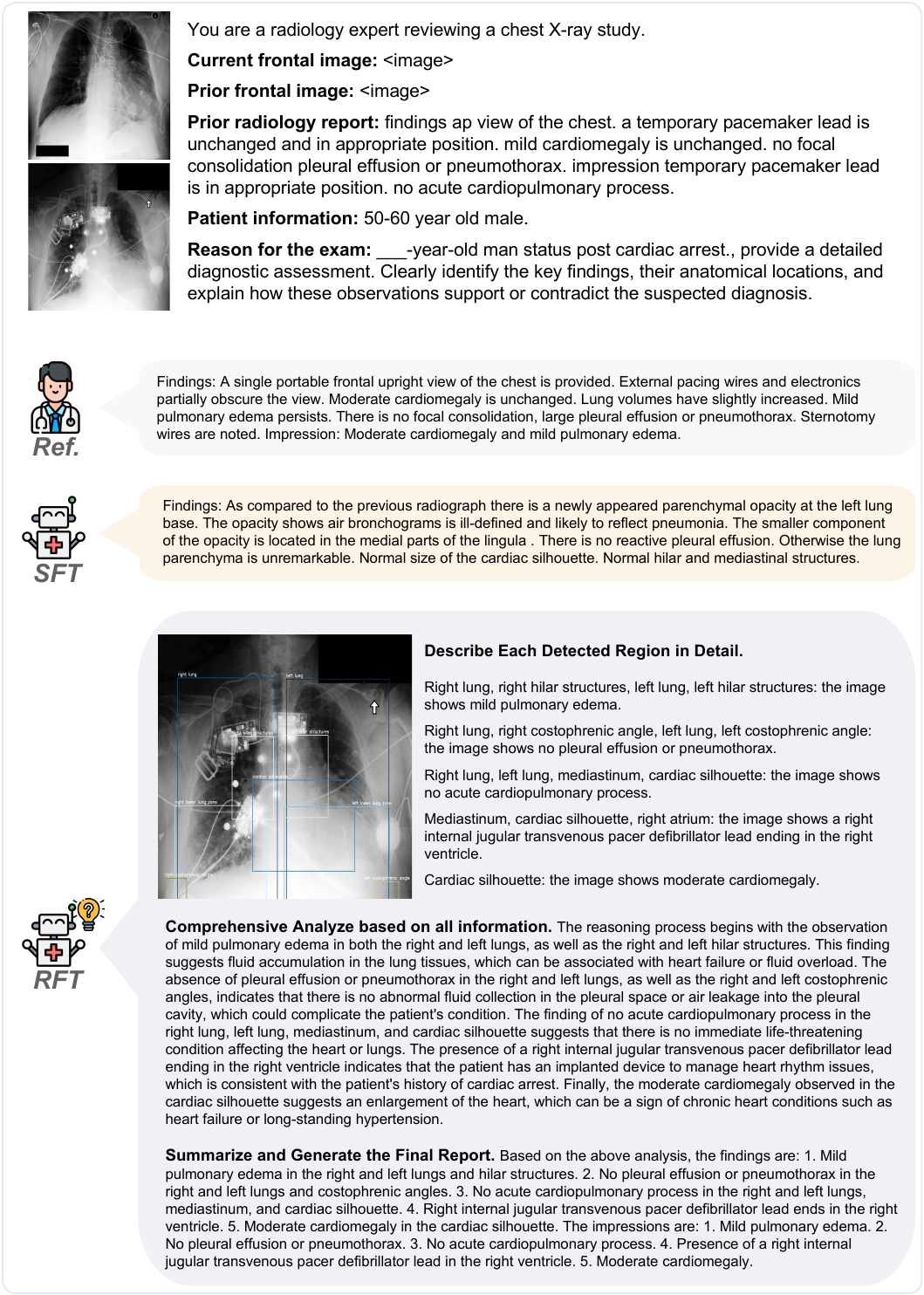}
    \caption{Representative example of trustworthy report generation with integrated visual and textual explainability.}
    \label{fig:Reasoning_result}
\end{figure}

\clearpage
\newpage

\begin{table}[ht]
\centering
\caption{Comparison of Methods Across Medical VQA Benchmarks with Best in Bold}
\label{tab:vqa}
\begin{adjustbox}{max width=\textwidth}
\begin{tabular}{llcccccc|c}
\toprule
\textbf{Dataset} & \textbf{Metric} & \textbf{RadFM} & \textbf{LLaVA-Med} & \textbf{MedFlamingo} & \textbf{InternVL} & \textbf{MedDr} & \textbf{MedRegA} & \textbf{Ours} \\
\midrule
\multirow{7}{*}{Slake-VQA}
 & BLEU-1     & 0.7522 & 0.6619 & 0.1087 & 0.3776 & 0.7648  & 0.8155 & \textbf{0.8844} \\
 & ClosedAcc  & 0.7408 & 0.3887 & 0.4648 & 0.1718 & 0.8338  & 0.8535 & \textbf{0.9239} \\
 & OpenRecall & 0.7726 & 0.8155 & 0.2719 & 0.5076 & 0.7418  & 0.8045 & \textbf{0.8774} \\
 & Recall     & 0.7620 & 0.6727 & 0.3280 & 0.3953 & 0.7726  & --     & \textbf{0.8930} \\
 & OpenAcc    & 0.7408 & 0.7819 & 0.2295 & 0.4703 & 0.7053  & --     & \textbf{0.8470} \\
 & F1         & 0.7598 & 0.6687 & 0.1595 & 0.3871 & 0.7756  & 0.8206 & \textbf{0.8917} \\
\midrule
\multirow{7}{*}{PathVQA}
 & BLEU-1     & 0.2489 & 0.4479 & 0.1027 & 0.0340 & 0.6143  & 0.5952 & \textbf{0.6280} \\
 & ClosedAcc  & 0.4886 & 0.5600 & 0.5739 & 0.0307 & 0.9021  & 0.8912 & \textbf{0.9071} \\
 & OpenRecall & 0.0249 & \textbf{0.3791} & 0.0664 & 0.0499  & 0.2800 & 0.3356 & 0.3649 \\
 & Recall     & 0.2570 & 0.4607 & 0.3202 & 0.0403 & 0.6192  & --     & \textbf{0.6368} \\
 & OpenAcc    & 0.0128 & \textbf{0.3487} & 0.0457 & 0.0353  & --     & --     & 0.3282 \\
 & F1         & 0.2524 & 0.4547 & 0.1522 & 0.0384 & 0.6215  & 0.6069 & \textbf{0.6344} \\
\midrule
\multirow{7}{*}{VQA-RAD}
 & BLEU-1     & 0.4864 & 0.3746 & 0.1423 & 0.2626 & 0.5962  & \textbf{0.6189} & 0.6083 \\
 & ClosedAcc  & 0.6534 & 0.2231 & 0.4741 & 0.2709 & 0.7888  & 0.7530 & \textbf{0.8048} \\
 & OpenRecall & 0.3683 & \textbf{0.6335} & 0.2573 & 0.3408 & 0.3745  & 0.4603 & 0.4507 \\
 & Recall     & 0.5270 & 0.3851 & 0.3758 & 0.3019 & 0.6051   & --     & \textbf{0.6478} \\
 & OpenAcc    & 0.3150 & \textbf{0.5950} & 0.1800 & 0.2750 & 0.3000   & --     & 0.3500 \\
 & F1         & 0.5005 & 0.3838 & 0.1998 & 0.2777 & 0.6110  & 0.6224 & \textbf{0.6282} \\
\bottomrule
\end{tabular}
\end{adjustbox}
\end{table}

\begin{table}[ht]
\centering
\caption{Performance Comparison on Single-Label and Multi-Label Benchmarks with Best in Bold}
\label{tab:sinlge_multi_label_diagnosis}
\begin{adjustbox}{max width=\textwidth}
\begin{tabular}{llcccccc|c}
\toprule
\textbf{Category} & \textbf{Dataset} & \textbf{RadFM} & \textbf{LLaVA-Med} & \textbf{MedFlamingo} & \textbf{InternVL} & \textbf{MedDr} & \textbf{MedRegA} & \textbf{Ours} \\
\midrule
\multirow{9}{*}{Single-Label}
 & OrganAMNIST     & 0.0123 & 0.1490 & 0.0478 & 0.0891 & 0.2066 & 0.2937 & \textbf{0.9356} \\
 & OrganCMNIST     & 0.0662 & 0.0750 & 0.0149 & 0.0462 & 0.0699 & 0.2422 & \textbf{0.8546} \\
 & OrganSMNIST     & 0.0649 & 0.0727 & 0.0132 & 0.0467 & 0.0868 & 0.1898 & \textbf{0.7138} \\
 & PathMNIST       & --     & 0.0834 & --     & 0.0685 & 0.2306 & --     & \textbf{0.9556} \\
 & OCTMNIST        & 0.1997 & 0.1305 & 0.1000 & 0.0929 & 0.5831 & 0.6446 & \textbf{0.8834} \\
 & PneumoniaMNIST  & 0.5865 & 0.5000 & 0.7131 & 0.2359 & 0.8734 & 0.8523 & \textbf{0.9400} \\
 & BreastMNIST     & 0.3526 & 0.3782 & 0.3077 & 0.0640 & 0.7179 & 0.7035 & \textbf{0.8329} \\
 & PAD-UFES-20     & 0.1381 & 0.1290 & 0.0259 & 0.0220 & 0.1409 & 0.2024 & \textbf{0.4746} \\
 & MURA            & 0.4950 & 0.3631 & 0.3430 & 0.0154 & 0.3781 & 0.6999 & \textbf{0.7938} \\
\midrule
\multirow{8}{*}{Multi-Label}
 & ChestMNIST      & 0.0491 & --     & --     & --     & 0.1339 & 0.1196 & \textbf{0.2386} \\
 & VinDr-CXR       & 0.0657 & 0.0236 & 0.0159 & 0.0225 & 0.0711 & 0.1189 & \textbf{0.3423} \\
 & VinDr-PCXR      & 0.0808 & 0.0192 & 0.0231 & 0.0220 & 0.0817 & 0.0633 & \textbf{0.1122} \\
 & VinDr-SpineXR   & 0.1667 & 0.0249 & 0.0847 & 0.0817 & 0.2682 & 0.3016 & \textbf{0.3852} \\
 & VinDr-Mammo     & 0.1389 & 0.0325 & 0.0085 & 0.0154 & 0.1935 & 0.0999 & \textbf{0.1621} \\
 & RFMiD           & 0.0554 & 0.0265 & 0.0153 & 0.0225 & 0.0189 & 0.0655 & \textbf{0.2160} \\
 & BRSET           & 0.0676 & 0.0793 & 0.0248 & 0.0817 & 0.0249 & 0.1128 & \textbf{0.3786} \\
\bottomrule
\end{tabular}
\end{adjustbox}
\end{table}

\begin{table}[ht]
\centering
\caption{Performance Comparison on Medical Report Generation Benchmarks with Best in Bold}
\label{tab:report_generation}
\begin{adjustbox}{max width=\textwidth}
\begin{tabular}{llcccccc|c}
\toprule
\textbf{Dataset} & \textbf{Metric} & \textbf{RadFM} & \textbf{LLaVA-Med} & \textbf{MedFlamingo} & \textbf{InternVL} & \textbf{MedDr} & \textbf{MedRegA} & \textbf{Ours} \\
\midrule
\multirow{3}{*}{MIMIC-CXR}
 & BLEU-4          & 0.0555 & 0.0096 & 0.0191 & 0.0133 & 0.0759 & \textbf{0.1030} & 0.0929 \\
 & ROUGE-L         & 0.2052 & 0.1390 & 0.1460 & 0.1250 & 0.2259 & 0.2440 & \textbf{0.2528} \\
 & CheXbert-F1@Pos    & -- & -- & -- & 0.2770 & 0.1030 & 0.4190 & \textbf{0.4790} \\
\midrule
\multirow{3}{*}{IU-Xray}
 & BLEU-4          & 0.1028 & 0.0068 & 0.0206 & 0.0156 & 0.1222 & \textbf{0.1480} & 0.1455 \\
 & ROUGE-L         & 0.2607 & 0.0996 & 0.1111 & 0.1218 & 0.2835 & 0.3040 & \textbf{0.3335} \\
 & CheXbert-F1@Pos    & -- & -- & -- & 0.2160 & 0.5647 & 0.5440 & \textbf{0.5690} \\
\bottomrule
\end{tabular}
\end{adjustbox}
\end{table}

\begin{table}[ht]
\centering
\caption{Performance Comparison on GMAI-MMbench with Best in Bold}
\label{tab:GMAI_MMbench}
\begin{adjustbox}{max width=\textwidth}
\begin{tabular}{llcccccc|c}
\toprule
\textbf{Category} & \textbf{Task} & \textbf{GPT-4O} & \textbf{LLaVA-Med} & \textbf{MedFlamingo} & \textbf{InternVL} & \textbf{MedDr} & \textbf{MedRegA} & \textbf{Ours} \\
\midrule
\multirow{19}{*}{GMAI}
 & Overall                         & 0.5396 & 0.1960 & 0.1164 & 0.4360 & 0.4369 & 0.4422 & \textbf{0.6460}  \\
 & Attribute Recognition           & 0.3832 & 0.2451 & 0.0667 & 0.5111 & 0.4120 & 0.3556 & \textbf{0.8755} \\
 & Blood Vessels Recognition       & 0.6101 & 0.1783 & 0.1014 & 0.4593 & 0.5070 & 0.4444 & \textbf{0.7743} \\
 & Bone                            & 0.5708 & 0.1708 & 0.0923 & 0.3771 & 0.3785 & 0.5371 & \textbf{0.8827} \\
 & Cell Recognition                & \textbf{0.4902} & 0.1986 & 0.1127 & 0.4609 & 0.2987 & 0.3043 & 0.4492 \\
 & Counting                        & 0.4662 & 0.1504 & 0.0662 & 0.3511 & 0.2827 & 0.2872 & \textbf{0.6346} \\
 & Disease Diagnosis               & 0.6145 & 0.1981 & 0.1343 & 0.5019 & \textbf{0.5253} & 0.5184 & 0.6530 \\
 & Image Quality Grading           & 0.4656 & 0.2024 & 0.1215 & 0.3600 & 0.3603 & 0.3200 & \textbf{0.7320} \\
 & Microorganism Recognition       & 0.5638 & 0.2151 & 0.0638 & 0.5481 & 0.3145 & 0.4593 & \textbf{0.8131} \\
 & Muscle                          & 0.3400 & 0.1320 & 0.0800 & 0.2400 & 0.296  & 0.3800 & \textbf{0.6000} \\
 & Nervous Tissue                  & \textbf{0.7525} & 0.1515 & 0.1818 & 0.625  & 0.4747 & 0.5250 & 0.6700 \\
 & Organ Recognition - Abdomen     & 0.5379 & 0.2042 & 0.0926 & 0.3061 & 0.3337 & 0.4286 & \textbf{0.6509} \\
 & Organ Recognition - Head \& Neck& 0.6947 & 0.2373 & 0.1827 & 0.4581 & 0.5133 & 0.5226 & \textbf{0.7329} \\
 & Organ Recognition - Pelvic      & 0.4867 & 0.1767 & 0.1100 & 0.4400 & 0.3267 & 0.2533 & \textbf{0.5333} \\
 & Organ Recognition - Thorax      & 0.6588 & 0.1965 & 0.1153 & 0.3882 & 0.4447 & 0.3882 & \textbf{0.7624} \\
 & Severity Grading                & 0.3393 & 0.2170 & 0.1216 & 0.3631 & 0.3514 & 0.3690 & \textbf{0.3847} \\
 & Surgeon Action Recognition      & 0.2288 & 0.1981 & 0.0519 & 0.2522 & 0.2519 & 0.2435 & \textbf{0.7154} \\
 & Surgical Instrument Recognition & 0.2951 & 0.1411 & 0.0847 & 0.2510 & 0.2558 & 0.2510 & \textbf{0.4076} \\
 & Surgical Workflow Recognition   & 0.3943 & 0.2086 & 0.1143 & 0.2571 & 0.3229 & 0.2857 & \textbf{0.5428} \\
\bottomrule
\end{tabular}
\end{adjustbox}
\end{table}

\begin{table}[ht]
\centering
\caption{Performance on OmniMedVQA Benchmark with Best in Bold with Best in Bold}
\label{tab:OmniMedVQA}
\begin{adjustbox}{max width=\textwidth}
\begin{tabular}{llcccccc|c}
\toprule
\textbf{Category} & \textbf{Modality} & \textbf{RadFM} & \textbf{LLaVA-Med} & \textbf{MedFlamingo} & \textbf{InternVL} & \textbf{MedDr} & \textbf{MedRegA} & \textbf{Ours} \\
\midrule
\multirow{8}{*}{OmniMedVQA}
 & CT                 & 0.3330 & 0.2530 & 0.3460 & 0.5963 & 0.6934 & 0.6564 & \textbf{0.8251} \\
 & Fundus Photography & 0.3500 & 0.4840 & 0.3330 & 0.7638 & 0.7501 & 0.7977 & \textbf{0.8099} \\
 & MRI                & 0.2200 & 0.3590 & 0.2750 & 0.7338 & \textbf{0.8162} & 0.7050 & 0.7866 \\
 & OCT                & 0.3130 & 0.4210 & 0.2600 & 0.7548 & \textbf{0.9139} & 0.8506 & 0.8450 \\  % Note: 9139 > 8506 > 8450
 & Dermatoscopy       & 0.3630 & 0.4520 & 0.2830 & 0.7289 & 0.7224 & 0.7898 & \textbf{0.8913} \\
 & Microscopy         & 0.2800 & 0.4400 & 0.2810 & 0.7761 & 0.6409 & 0.7125 & \textbf{0.8234} \\
 & X-Ray              & 0.3150 & 0.3170 & 0.3010 & 0.8305 & 0.7900 & 0.7835 & \textbf{0.9142} \\
 & Ultrasound         & 0.2610 & \textbf{0.8370} & 0.3320 & 0.6163 & 0.4177 & 0.4661 & 0.8264 \\
\bottomrule
\end{tabular}
\end{adjustbox}
\end{table}

\begin{table}[ht]
\centering
\caption{Performance on Region Recognition and Lesion Detection Tasks with Best in Bold}
\label{tab:region_rec_lesion_det}
\begin{adjustbox}{max width=\textwidth}
\begin{tabular}{llcc|c}
\toprule
\textbf{Task} & \textbf{Metric} & \textbf{MedRegA} & \textbf{InternVL} & \textbf{Ours} \\
\midrule
\multirow{6}{*}{Region Recognition}
 & Head and Neck & 0.4972 & 0.1340 & \textbf{0.7327} \\
 & Chest         & 0.3487 & 0.0460 & \textbf{0.7186} \\
 & Abdomen       & 0.4528 & 0.0720 & \textbf{0.7273} \\
 & Skeleton      & \textbf{0.5921} & 0.0000 & 0.4713 \\
 & Pathology     & 0.2222 & 0.0000 & \textbf{0.4828} \\
 & Overall       & 0.4226 & 0.0504 & \textbf{0.6265} \\ 
\midrule
\multirow{6}{*}{Lesion Detection}
 & Head and Neck & 0.4378 & -- & \textbf{0.9777} \\
 & Chest         & 0.3581 & -- & \textbf{0.9419} \\
 & Abdomen       & 0.4664 & -- & \textbf{0.8792} \\
 & Skeleton      & 0.3103 & -- & \textbf{0.9013} \\
 & Pathology     & 0.1121 & -- & \textbf{0.8095} \\
 & Overall       & 0.3369 & -- & \textbf{0.9019} \\
\bottomrule
\end{tabular}
\end{adjustbox}
\end{table}

\begin{table}[ht]
\centering
\caption{Performance on MedGRIT Region VQA Benchmark with Best in Bold}
\label{tab:MedGRIT}
\begin{adjustbox}{max width=\textwidth}
\begin{tabular}{llccccc}
\toprule
\textbf{Task} & \textbf{SubTask} & \textbf{BiRD} & \textbf{LLaVA} & \textbf{MedRegA} & \textbf{InternVL} & \textbf{Ours} \\
\midrule
\multirow{4}{*}{Region VQA}
 & VG         & 0.5392 & 0.0000 & 0.0000 & 0.0000 & \textbf{0.7411} \\
 & ROC          & 0.6533 & 0.0275 & 0.0724 & 0.1341 & \textbf{0.8857} \\
 & RC  & 0.5523 & 0.0818 & 0.0569 & 0.1065 & \textbf{0.6908} \\
 & MIA & 0.5217 & 0.1120 & 0.0671 & 0.0565 & \textbf{0.6156} \\
\bottomrule
\end{tabular}
\end{adjustbox}
\end{table}

\begin{table}[ht]
\centering
\caption{Audio-Input Evaluation Results Across Diverse Benchmarks with Best in Bold}
\label{tab:audio}
\begin{adjustbox}{max width=\textwidth}
\begin{tabular}{llccc}
\toprule
\textbf{Task} & \textbf{Dataset} & \textbf{Intern\_omni} & \textbf{Ours\_audio} & \textbf{Ours\_text} \\
\midrule
\multirow{3}{*}{VQA}
 & Slake         & 0.4680 & 0.6700 & \textbf{0.8917} \\
 & VQA-Rad       & 0.4350 & 0.4820 & \textbf{0.6282} \\
 & PathVQA       & 0.2690 & 0.5230 & \textbf{0.6344} \\
\midrule
\multirow{4}{*}{Multi-label Diagnosis}
 & VinDr-CXR     & 0.0430 & 0.2460 & \textbf{0.3423} \\
 & VinDr-PCXR    & 0.0210 & 0.0840 & \textbf{0.1122} \\
 & VinDr-SpineXR & 0.0440 & 0.3640 & \textbf{0.3852} \\
 & VinDr-Mammo   & 0.0140 & 0.1300 & \textbf{0.1621} \\
\midrule
\multirow{5}{*}{Single-label Diagnosis}
 & OrganAMNIST        & 0.0680 & \textbf{0.9400} & 0.9356 \\
 & OrganCMNIST        & 0.0250 & \textbf{0.9000} & 0.8546 \\
 & OrganSMNIST        & 0.0210 & \textbf{0.7490} & 0.7138 \\
 & BloodMNIST         & 0.0800 & 0.6030 & \textbf{0.8329} \\
 & BreastMNIST        & 0.1020 & 0.6930 & \textbf{0.8329} \\
\bottomrule
\end{tabular}
\end{adjustbox}
\end{table}

\begin{table}[ht]
\centering
\caption{Comparison of Region Recognition and Lesion Detection on Internal vs External Test Sets with Best in Bold}
\label{tab:external_regionrec_lesiondet}
\begin{adjustbox}{max width=\textwidth}
\begin{tabular}{llcc}
\toprule
\textbf{Disease} & \textbf{Task} & \textbf{Internal} & \textbf{External} \\
\midrule
\multirow{2}{*}{Brain Tumor}
 & Region Recognition & \textbf{0.9990} & 0.7150 \\
 & Lesion Detection   & \textbf{0.7970} & 0.6560 \\
\midrule
\multirow{2}{*}{Acoustic Neuroma}
 & Region Recognition & \textbf{0.9990} & 0.8860 \\
 & Lesion Detection   & \textbf{0.8480} & 0.6290 \\
\midrule
\multirow{2}{*}{Cerebral Hemorrhage}
 & Region Recognition & 0.9300 & \textbf{0.9990} \\
 & Lesion Detection   & \textbf{0.6710} & 0.4980 \\
\midrule
\multirow{2}{*}{Lung Cancer}
 & Region Recognition & \textbf{0.9990} & 0.9810 \\
 & Lesion Detection   & \textbf{0.6340} & 0.3710 \\
\midrule
\multirow{2}{*}{Pulmonary Embolism}
 & Region Recognition & \textbf{0.9990} & 0.9490 \\
 & Lesion Detection   & \textbf{0.5240} & 0.0850 \\
\midrule
\multirow{2}{*}{Liver Cancer}
 & Region Recognition & \textbf{0.9960} & 0.9800 \\
 & Lesion Detection   & 0.4470 & \textbf{0.4880} \\
\midrule
\multirow{2}{*}{Pancreatic Cancer}
 & Region Recognition & 0.9360 & \textbf{0.9560} \\
 & Lesion Detection   & \textbf{0.5540} & 0.4120 \\
\midrule
\multirow{2}{*}{Kidney Cancer}
 & Region Recognition & \textbf{0.9760} & 0.6390 \\
 & Lesion Detection   & \textbf{0.7800} & 0.5480 \\
\midrule
\multirow{2}{*}{Colon Cancer}
 & Region Recognition & \textbf{0.1540} & 0.1070 \\
 & Lesion Detection   & \textbf{0.6880} & 0.4240 \\
\midrule
\multirow{2}{*}{Prostatic Cancer}
 & Region Recognition & \textbf{0.8700} & 0.4520 \\
 & Lesion Detection   & \textbf{0.3760} & 0.1520 \\
\midrule
\multirow{2}{*}{Bladder Cancer}
 & Region Recognition & \textbf{0.9600} & 0.7890 \\
 & Lesion Detection   & \textbf{0.5150} & 0.4740 \\
\bottomrule
\end{tabular}
\end{adjustbox}
\end{table}

\begin{table}[ht]
\centering
\caption{Region VQA external validation results. Scores represent \textit{mean} \( \pm \) \textit{standard deviation} across four evaluation metrics rated by large language models (GPT-4o). All metrics are scored out of 10, with higher values indicating better performance.}
\label{tab:external_vqa}
\begin{adjustbox}{max width=\textwidth}
\begin{tabular}{lccccc|c}
\toprule
\textbf{Metric} & \textbf{MedDr} & \textbf{MedRegA} & \textbf{LLaVA-Med} & \textbf{MedFlamingo} & \textbf{RadFM} & \textbf{Ours} \\
\midrule
Consistency          & \(6.82 \pm 1.64\) & \(8.11 \pm 1.35\) & \(8.01 \pm 1.54\) & \(3.20 \pm 1.23\) & \(4.72 \pm 2.17\) & \(\mathbf{8.67 \pm 0.65}\) \\
Clinical Relevance   & \(6.56 \pm 1.60\) & \(7.76 \pm 1.20\) & \(7.83 \pm 1.35\) & \(3.05 \pm 1.21\) & \(4.55 \pm 2.18\) & \(\mathbf{8.50 \pm 0.59}\) \\
Accuracy             & \(6.66 \pm 1.74\) & \(8.07 \pm 1.41\) & \(7.82 \pm 1.66\) & \(3.42 \pm 1.32\) & \(4.46 \pm 2.23\) & \(\mathbf{8.65 \pm 0.68}\) \\
\midrule 
Mean        & \(7.98 \pm 1.29\) & \(6.68 \pm 1.64\) & \(7.89 \pm 1.49\) & \(3.22 \pm 1.21\) & \(4.58 \pm 2.18\) & \(\mathbf{8.61 \pm 0.60}\) \\
\bottomrule
\end{tabular}
\end{adjustbox}
\end{table}

\end{document}